\documentclass[12pt]{article}
\usepackage[a4paper,left=1in,top=1in,right=1in,bottom=1in,nohead]{geometry}
\usepackage[onehalfspacing]{setspace}
\usepackage[american]{babel}
\usepackage{lmodern}
\usepackage[utf8]{inputenc} 
\usepackage[T1]{fontenc}    
\usepackage{url}
\usepackage{booktabs} 
\usepackage{amsfonts}
\usepackage{xcolor}
\usepackage{natbib}
\usepackage[pdftex]{graphicx}
\usepackage[export]{adjustbox}
\usepackage{amsmath}
\usepackage{colortbl}
\usepackage[hidelinks]{hyperref}
\usepackage{gb4e}
\noautomath 

\title{Left-right asymmetry in predicting brain activity from LLMs' representations emerges with their formal linguistic competence}

\author{%
Laurent Bonnasse-Gahot\\
Centre d'Analyse et de Mathématique Sociales\\
CNRS, EHESS, Paris, France\\
\texttt{lbg@ehess.fr}\\[12pt]
Christophe Pallier\\
Cognitive Neuroimaging Unit\\
CNRS, INSERM, CEA, Neurospin Center, 91191 Gif-sur-Yvette, France\\
\texttt{christophe@pallier.org}
$\,$\\[20pt]
}

\date{}

\begin{document}

\begin{singlespacing}
\maketitle
\end{singlespacing}

\begin{abstract}
When humans and large language models (LLMs) process the same text, activations in the LLMs correlate with brain activity measured, e.g., with functional magnetic resonance imaging (fMRI). Moreover, it has been shown that, as the training of an LLM progresses, the performance in predicting brain activity from its internal activations improves more in the left hemisphere than in the right one. The aim of the present work is to understand which kind of competence acquired by the LLMs underlies the emergence of this left-right asymmetry. Using the OLMo-2 7B language model at various training checkpoints and fMRI data from English participants, we compare the evolution of the left-right asymmetry in the correlation between brain activity and model predictions alongside performance on several benchmarks. We observe that the asymmetry co-emerges with the formal linguistic abilities of the LLM. These abilities are demonstrated in two ways: by the model's capacity to assign a higher probability to an acceptable sentence than to a grammatically unacceptable one within a minimal contrasting pair, and by its ability to produce well-formed text. By contrast, the left-right asymmetry does not align with the performance on arithmetic or Dyck language tasks; nor with text-based tasks involving world knowledge and reasoning. We generalize these results to another family of LLMs (Pythia) and two other languages, French and Chinese. Our observations indicate that the left-right asymmetry in brain predictivity matches the progress in formal linguistic competence.
\end{abstract}

\section{Introduction}
The success of large language models (LLMs) in natural language processing tasks has generated a lot of interest in understanding their internal representations and their alignment with human brain activity. Brain activations, measured with functional magnetic resonance, magnetoencephalography, or electrocorticography, in humans listening to or reading a text can be predicted from the internal activity of LLMs fed with the same text \citep{jain2018incorporating,toneva2019interpreting,schrimpf2021neural,caucheteux2022brains,goldstein2022shared,pasquiou_information-restricted_2023,antonello2024scaling,hosseini2024artificial}. In the encoding approach, brain activations are regressed on the hidden neural activations from an LLM and the resulting model is used to compute brain-predicitity scores, that is, cross-validated correlations at each voxel \citep[see, e.g.,][for a review]{dupre2025voxelwise}.

Early studies \citep{huth2016natural,jain2018incorporating,caucheteux2021disentangling,pasquiou_information-restricted_2023} reported brain correlation maps that were very symmetrical, with similar values in both hemispheres, an odd finding given the evidence for left hemispheric dominance for language. For instance, in their seminal paper, \citet{huth2016natural} noted that ``One striking aspect of our atlas is that the distribution of semantically selective areas is relatively symmetrical across the two cerebral hemispheres. This finding is inconsistent with human lesion studies that support the idea that semantic representation is lateralized to the left hemisphere.'' The question of lateralization is foundational to the field of the neurobiology of language since the seminal reports by Dax and Broca linking  aphasia to left hemispheric lesions \citep{dax_lesions_1865,broca1865siege}. It has remained an important theme over decades \citep{penfield_speech_1959,wada_intracarotid_1960,gazzaniga_language_1967,witelson1973left,binder_determination_1996,just1996brain}, up until the recent years \citep{bradshaw2017measuring,tzourio-mazoyer_multi-factorial_2017,hausmann_language_2019,malik2022investigation,ozernov2026precision}. Although functional imaging studies have established that both hemispheres are involved in language comprehension, they have also confirmed that the left hemisphere is dominant, e.g. show stronger and more widespread activations during speech processing \citep[see e.g.][]{hickok2007}.

Recently, we showed that the symmetrical results reported with small, first-generation LLMs, disappeared with larger and higher-performing models \citep{bonnasse-gahot_fmri_2024}. More precisely, brain predictivity maps exhibited an increasing left-right hemispheric asymmetry when LLMs increased in number of parameters and in performance on natural language processing tasks. Furthermore, this left-right asymmetry also emerged for a given LLM alongside its training. The relationship between amount of training and left-right asymmetry showed a phase transition profile that is reminiscent of those that have been observed in LLMs' performance on several benchmarks \citep{chen2023sudden}.

The present work aims at understanding what competence acquired during training drives the emergence of the left-right asymmetry in brain predictivity. We conduct a series of experiments designed to track the evolution of linguistic and non-linguistic capabilities of LLMs as a function of training progression, and we study their relationship to left-right asymmetry in brain correlation. In our initial experiment, we systematically investigate how the performance of an LLM on a set of carefully constructed benchmarks evolves with training. This set includes two linguistic benchmarks (BLiMP, \citealp{warstadt2020blimp} and Zorro, \citealp{huebner2021babyberta}) and two non-linguistic benchmarks (specifically, Arithmetic and Dyck language tasks), all designed as minimal-pair tasks to isolate specific competencies. Our analyses reveal a striking correlation: as training progresses, the emergence of the left-right dominance in the correlation between brain activity and model predictions closely mirrors the improvement in performance on the linguistic benchmarks, but not on the non-linguistic benchmarks. 

BLiMP and Zorro essentially assess formal linguistic competence (knowledge of linguistic rules and patterns). In a follow-up experiment, focusing on text-based tasks, we contrast formal linguistic competence and functional linguistic competence (understanding and using language in the world), a distinction proposed by \citet{mahowald2024dissociating}. To assess the model's formal competence beyond BLiMP and Zorro, we evaluate the linguistic acceptability of texts generated by the model at different checkpoints during training. To assess functional competence, we test the LLM on several conceptual and reasoning benchmarks:  GSM8K \citep{cobbe2021gsm8k}, ARC \citep{clark2018think}, Hellaswag \citep{zellers2019hellaswag}, and WinoGrande \citep{sakaguchi2021winogrande}. The results show that the phase transition of the left-right asymmetry in brain-predictivity aligns with the linguistic acceptability scores, and not with any of the four benchmarks which assess functional competence.

The results just described are based on OLMo-2 7B model \citep{olmo20242olmo2furious}, a recent model for which training checkpoints are available. We show that these results generalize to other models, namely the 2.8b and the 6.9b models from the Pythia family \citep{biderman2023pythia}. We then test two other languages: French and Chinese using datasets from the \textit{Le Petit Prince} project. We replicate in these two languages the finding that the left-right asymmetry aligns better with tests of formal competence than with tests of functional competence. 

Collectively, these results support the hypothesis that the emergence of the left-right asymmetry in LLMs' brain predictivity is a direct reflection of their acquisition of formal linguistic competence. 

\section{Materials and Methods}

\subsection{Brain Imaging Data}

The experiments reported in this paper rely on functional magnetic resonance data provided by the multilingual project \textit{Le Petit Prince}, in which English, French, and Mandarin Chinese speakers were scanned while listening for a bit more that an hour and a half to an audiobook of \textit{The Little Prince} \citep{li2022petit}. The presentation of the audiobook was split into 9 parts of approximately equal duration, during which functional images of the full brain were acquired every 2~s. After spatially normalizing these images into a common space and resampling them at 4$\times$4$\times$4mm, we average the time-series across participants (high-pass filtered with a cut-off of 128~s and standardized in each voxel), to obtain an average English subject (from all 49 English participants). The same procedure was applied for the replication experiments in the two other languages, yielding an average French subject (from all 28 French participants) and an average Chinese subject (from all 35 Chinese participants). These three average subjects are available at \url{https://github.com/l-bg/llms\_brain\_lateralization}.

\subsection{Language models}
\label{sec:model}
The main large language model used in this study is the 7B-parameter version of OLMo-2~\citep{olmo20242olmo2furious}, released by Allen AI, and available at \url{https://huggingface.co/allenai/OLMo-2-1124-7B}. As far as we know, this is the best open-weight model under 10B parameters that releases a sufficient number of training checkpoints that allow to study the evolution of the performance of an LLM during training. The model has 32 layers and a hidden size of 4096. We consider 10 checkpoints from the base model: the first checkpoint available, after training on 1B tokens; the final checkpoint of their Stage 1 pretraining phase, which is the main part of their pretraining, corresponding to 1 epoch on the OLMo-Mix-1124 dataset (approximately 4T tokens); and 8 intermediate checkpoints log-spaced between these two extremes (using the closest available checkpoint). The main experiments in this study are based on this model. To check that the results are not specific to it, we run additional experiments using Pythia 2.8b and Pythia 6.9b \citep{biderman2023pythia}.

\subsection{Brain-predictivity scores}
For each voxel, we compute a brain-predictivity score that quantifies how well we can predict brain activity from the activations of the large language model, using the pipeline available at \url{https://github.com/l-bg/llms\_brain\_lateralization} and described in detail in \citet{bonnasse-gahot_fmri_2024}. In brief, this pipeline follows a standard nested cross-validation approach where the fMRI time series in a given voxel is fit with a linear model, regularized using ridge regression, on the activations obtained at a given layer of the LLM. As the fMRI data consist of 9 runs, cross-validation was achieved by fitting the model on 8 of the functional runs and testing it on the left-out run, in a roving fashion. 

In order to mimic the BOLD response, these artificial activations are convolved with the Glover haemodynamic response function \citep{glover1999deconvolution}. The brain-predictivity score associated with a given voxel is the maximum correlation associated with the best layer. Finally, left and right hemisphere brain-predictivity scores are obtained by averaging correlations from voxels located in the left and right hemispheres respectively. Unless otherwise indicated, we included only the 25\% vowels with the highest inter-subject correlations, to focus on brain areas where the signal is most reliable across participants. The masks corresponding to these voxels are displayed on Fig.~\ref{fig:masks}, separately for each language.

\subsection{Experiments}

\subsubsection*{Experiment 1: Minimal-pair benchmarks}

Minimal-pair benchmarks provide pairs of strings that differ minimally, one of which is ``acceptable'' in a language, while the other is not. As a language model assigns probabilities to strings, one can assess whether it would classify the acceptable string as more probable as the non-acceptable one. More precisely, given a string of words as the context, a causal model provides the probability distribution of the next word over the vocabulary of a language. The log probability of a sentence is then computed as the sum of the log probabilities of each word of the sentence. For a given set of sentences, the overall accuracy is calculated as the proportion (reported as a number between 0 and 1) of times the model correctly assigns a higher probability to the correct or most acceptable sentence.

We considered four minimal-pair benchmarks: two linguistic ones, BLiMP \citep{warstadt2020blimp} and Zorro \citep{huebner2021babyberta}, and two non-linguistic ones: Arithmetic and Dyck, specifically designed for this experiment. Here is a detailed description of these four benchmarks.

\noindent$\bullet$ \textbf{BLiMP}~\citep{warstadt2020blimp} provides 67,000 minimal pairs of English sentences, grouped into 67 paradigms of 1,000 pairs each, isolating specific phenomena in syntax, morphology, or semantics \citep[see][Table 4, for examples of each of these phenomena]{warstadt2020blimp}. Here is an example (the asterisk denotes the ungrammatical string), from the \textit{left branch island simple question} dataset:
\begin{exe}
\ex 
\begin{xlist}
\ex Whose hat should Tonya wear?
\ex * Whose should Tonya wear hat?
\end{xlist}
\end{exe}

\noindent$\bullet$ \textbf{Zorro}~\citep{huebner2021babyberta} is similar to BLiMP but uses a restricted vocabulary assumed to be known by a 6-year-old English child. Data consist of 22 files containing 4,000 sentences each (\textit{i.e.} 2,000 minimal pairs). Here is an example from the \textit{agreement subject verb across relative clause} paradigm:
\begin{exe}
\ex
\begin{xlist}
\ex The book that I like is poor.
\ex * The books that I like is poor.
\end{xlist}
\end{exe}

\noindent$\bullet$ The \textbf{Arithmetic} benchmark consists of an `addition' subtask and a `multiplication' subtask. Each subtask involves 2048 pairs of statements, one correct and one incorrect. The `addition' task considers statements of the form $x + y = z$, where $x$ and $y$ are randomly chosen between 0 and 1000. In the correct statement, $z$ is indeed the sum of $x$ and $y$, whereas in the incorrect version, an error randomly drawn from the set $[-10, -2, -1, 1, 2, 10]$ is added to the actual sum. In the `multiplication' task, statements are of the the form $x \times y = z$, where $x$ and $y$ are randomly chosen between 0 and 100. In the correct one, $z$ is the product of $x$ and $y$, whereas in the incorrect one, as for the previous `addition' task, we add to the product an error randomly drawn from the set $[-10, -2, -1, 1, 2, 10]$. The final accuracy is the mean accuracy over these two addition and multiplication tasks. Here is an example of a minimal pair:
\begin{exe}
 \ex 
 \begin{xlist}
 \ex 36 $\times$ 41 = 1476
 \ex * 36 $\times$ 41 = 1486
 \end{xlist}
\end{exe}

\noindent$\bullet$ The \textbf{Dyck} benchmark consists of three sub-benchmarks, based on the Dyck-1, Dyck-2, and Dyck-3 languages, which are formal languages that describe the balanced nesting of opening and closing parentheses (or other types of brackets). Here, Dyck-1 language involves the open and close parentheses `(' and `)', Dyck-2 uses parentheses and square brackets `(', `[', `]' and `)', and Dyck-3 parentheses, square brackets and curly brackets `(', `[', `\{', `]', `)' and `\}'. For each subtask, we randomly generate 1024 minimal pairs of sentences of length 32. For a given pair, the correct version is well-parenthesized, whereas we introduce errors in the incorrect version by randomly permuting two neighboring elements in the second half of the sentence, so that the two sentences share the same beginning and the same elements overall. Below is an example of such a minimal pair from the Dyck-3 benchmark. The final accuracy is the average of the accuracy on these three subtasks.
\begin{exe}
\ex 
\begin{xlist}
\ex \makebox[\textwidth][s]{\texttt{\small (\,(\,)\,[\,]\,)\,(\,)\,\{\,[\,]\,\}\,\{\,\}\,\{\,\}\,\{\,(\,)\,\}\,(\,)\,[\,]\,(\,\{\,\{\,\}\,\}\,)\,[\,]}}
\ex * \makebox[\textwidth][s]{\texttt{\small (\,(\,)\,[\,]\,)\,(\,)\,\{\,[\,]\,\}\,\{\,\}\,\{\,\}\,\{\,)\,(\,\}\,(\,[\,)\,]\,(\,\{\,\{\,\}\,\}\,[\,)\,]}}
\end{xlist}
\end{exe}

\subsubsection*{Experiment 2: Further tests of formal and functional competence}

While the first experiment aims at evaluating the language model in different domains with the same technique (the minimal-pair approach), experiment 2 further evaluates the linguistic competences of the model, with six new tests. The first is a test devised by us, which evaluates the linguistic acceptability of texts generated by the model. Like Zorro and BLiMP, it essentially evaluates the formal linguistic competence of the  model, but using a very different approach (text generation rather than sentence probability evaluation). The other tests are off-the-shelves evaluation benchmarks that assess the functional competence of the model in different domains: world knowledge, general and mathematical reasoning, and pronoun resolution \citep{zellers2019hellaswag,clark2018think,cobbe2021gsm8k,sakaguchi2021winogrande}. Let us describe these tests in detail.

\noindent$\bullet$ \textbf{Linguistic acceptability of text generations}: We assess the evolution during training of the linguistic acceptability of texts generated by the LLM. For each checkpoint during training, the LLM is asked to generate a continuation, between 192 and 256 tokens, from one of the following ten prompts: ``Why not'', ``Are you'', ``This is'', ``Alice was'', ``Bob went'', ``The thing'', ``Yet the'', ``A blue'', ``I wish'', ``Once upon''. Texts are generated five times for each prompt, each time with a different initial seed. Section~\ref{sec:supp_llm_gen} in Supplementary Materials provides samples for one prompt and a given seed, for all 10 checkpoints. All generated texts will be available on the GitHub page of the project. 
In order to automatically evaluate the acceptability of the generated texts, we use another LLM that was fine-tuned to output the linguistic acceptability of a sentence. This latter model is a version of DeBERTa-v3-large \citep{he2023debertav} fine-tuned on the CoLA dataset (model available on the Hugging Face hub at \url{https://huggingface.co/yiiino/deberta-v3-large-cola}). The Corpus of Linguistic Acceptability \citep[CoLA;][]{warstadt2019neural} is a widely used benchmark dataset for evaluating the ability of natural language processing models to judge the grammatical acceptability of English sentences. It consists of more than 10,000 English sentences labeled as either grammatical or ungrammatical \citep[see][Table 3, for samples]{warstadt2019neural}. The generated text is first split into sentences (using the \texttt{sent\_tokenize} function from \texttt{nltk} Python package, \citealp{bird2009natural}), then each sentence is fed into this fine-tuned LLM. The final linguistic acceptability score of the text is the mean score over all sentences in the text.

\noindent$\bullet$ \textbf{GSM8K}~\citep{cobbe2021gsm8k} aims at evaluating mathematical reasoning capabilities and consists of over 8,000 grade school math problems. Here is one of them: \emph{Kira bought 3 apples, 5 bananas and 6 oranges at the grocery store. Lola ate 2 pieces of the fruit. How many pieces are left?}

\noindent$\bullet$ \textbf{Hellaswag}~\citep{zellers2019hellaswag} is a completion test that assesses commonsense natural language inference. Given an event description, the language model must select the most likely followup among four choices. The 10,000 sentences were created to be very easy for humans but difficult for Natural Language Processing systems (that existed around the publication date). Here is an example. Given the context: \emph{A woman sits down at a piano. She...}, the model must select the most probable continuation among: a) \emph{sets her fingers on the keys} b) \emph{begins to eat a sandwich} c) \emph{stands up and walks to the kitchen} d) \emph{starts to paint the piano keys blue}.

\noindent$\bullet$ \textbf{ARC}~\citep{clark2018think} provides a question set which contains 7,787 natural, grade-school science questions (authored for human tests), assessing knowledge and reasoning according to the authors. The ARC question set is partitioned into an \textbf{Easy Set} and a \textbf{Challenge Set}. Here is an example of a question from ARC Easy: \emph{Which state of matter has no definite volume and no definite shape? a) gas (b) liquid c) solid}; and one example from ARC Challenge: \emph{George wants to warm his hands quickly by rubbing them. Which skin surface will produce the most heat?
a) dry palms
b) wet palms
c) palms covered with oil
d) palms covered with lotion.}

\noindent$\bullet$ \textbf{WinoGrande}~\citep{sakaguchi2021winogrande} is a dataset of 44,000 binary-choice challenges that aims at testing the ability of a language model to solve pronoun resolution problems (the ``Winograd Schema Challenge''). For example, given the sentence \emph{The trophy doesn’t fit into the brown suitcase because it is too large} or the sentence \emph{The trophy doesn’t fit into the brown suitcase because it is too small}, the model must decide which noun  the pronoun \emph{it} refers to (\emph{trophy} or \emph{suitcase}). 

\subsubsection*{Experiment 3: Replication with Pythia models}
In order to check that the results are not specific to OLMo-2-1124-7B, we replicate experiment 1, evaluating brain correlations and performance on the four minimal-pair benchmarks (BLiMP, Zorro, Arithmetics and Dyck) using two other models: Pythia-2.8b and Pythia-6.9b \citep{biderman2023pythia}. We consider 10 checkpoints during training, equally log-spaced, from step 16 (about 30M tokens) to step 143000 (about 300B tokens, the last step available of the pretraining phase).

\subsubsection*{Experiment 4: Replication in two other languages (French and Chinese)}
In this experiment, we applied a similar approach to the French text and the average French fMRI subject, and to the Chinese text and the average Chinese fMRI subject, still using the OLMo-2 model. Given that this model was mostly trained on English content~\citep{olmo20242olmo2furious}, with occasional texts from other languages, we expect its linguistic competence in French and Chinese to lag behind that in English. For French, we assess the formal linguistic competence using the fr-grammar task and the functional competence using the French Hellaswag (both tasks come from the FrenchBench benchmark; \citealp{faysse2025croissantllm}). For Chinese, formal competence is assessed with the ZhoBLiMP~\citep{liu2024systematic} benchmark, and functional competence with the CMMLU one (Chinese Massive Multitask Language Understanding; \citealp{li2024cmmlu}). We then compare the evolution during training of the performance on these tests with the respective left-right asymmetries in brain predictivity.

\subsubsection*{Experiment 5: Right-left hemispheric asymmetry in the cerebellum} 

Although the cerebellum's involvement in language processing has been known for some time \citep[see][for reviews]{murdoch2010cerebellum,highnam2011language}, there is a recent surge of interest around this topic \citep[see e.g.][]{lebel2023seat,fiez2024small}. In this vein, \citet{casto2025cerebellar} show that a region in the right cerebellar hemisphere is highly selective for language, both during comprehension and production, mirroring the neocortical language network located in the left hemisphere. This motivates us to check whether there is a \emph{right-left} asymmetry (as opposed to a left-right one) in predicting brain activity in the cerebellum from the internal representations of a LLM, and whether this right-left asymmetry also emerges along with the acquisition of formal linguistic competence. The masks used for the left and right cerebellum are displayed on Supplementary Fig.~\ref{fig:cerebellum_mask}.

\section{Results}

\subsection{Experiment 1: Minimal-pair benchmarks}

\begin{figure}
\centering
\includegraphics[width=\linewidth]{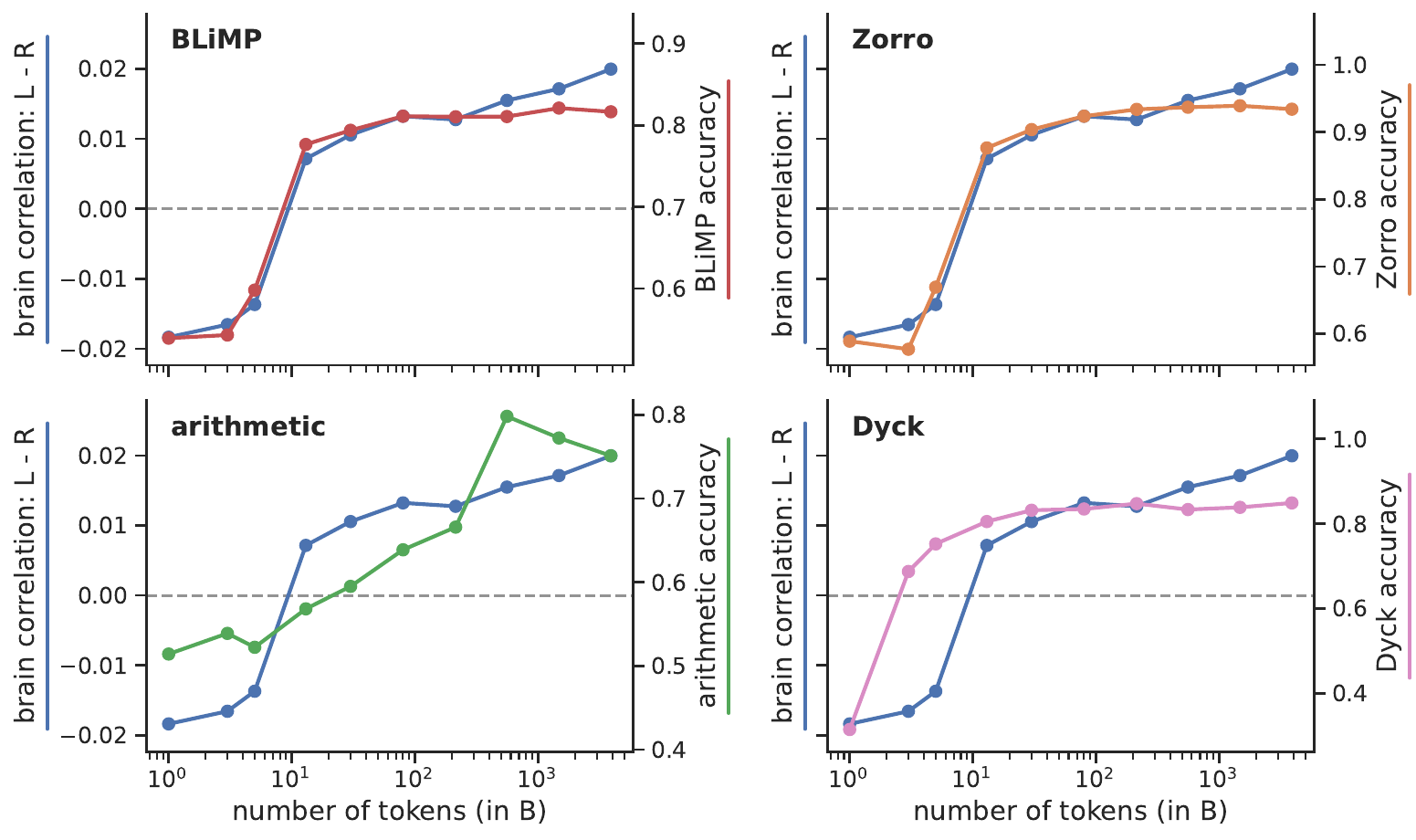} 
\caption{\textbf{Phase transitions during training: minimal-pair benchmarks.} Each panel displays the left-right hemispheric asymmetry in brain correlation (blue curve, repeated across panels) and the performance on a given test, as a function of the number of tokens seen during training (on a log scale). The top panels show the performance on the linguistic tests, BLiMP and Zorro, and the bottom panels show the performance on the non-linguistic tests, Arithmetic and Dyck. Model used: OLMo-2-1124-7B. Brain correlations are computed on the 25\% most reliable voxels. To help compare the transitions, the benchmarks curves were scaled along the y-axis to match the left-right asymmetry curve, by minimizing the mean sum of absolute values of orthogonal deviations \citep{nyquist1988least}.}
\label{fig:olmo-2_rv_acc}
\end{figure}

Fig.~\ref{fig:olmo-2_rv_acc} shows the evolution during training of the left-right asymmetry in  brain predictivity, computed with OLMo-2 7B, and the performance of this model on the four minimal-pair benchmarks (BLiMP, Zorro, Arithmetic and Dyck). 

First, a phase transition, that is, an abrupt change, occurs for the left-right asymmetry in brain predictivity (blue curve). The brain correlation in the left hemisphere becomes stronger than in the right as the model is trained on more tokens, reproducing in more detail the behavior reported in \citet[][Fig.~B10]{bonnasse-gahot_fmri_2024}. 
As for the performance on the four minimal-pair tests, the BLiMP and Zorro benchmarks (top panels) show a phase transition in the same interval as the left-right asymmetry, while the scores on the non-linguistics tests, Arithmetic and Dyck (bottom panels), do not follow the same pattern. The amount of training where formal linguistic abilities emerge is around 10B tokens, consistent with what \citet{tigges2024llm} reported for the Pythia family (see also our own results with Pythia below, section~\ref{sec:pythiaresults}). 

We further focus on the sub-tasks of BLiMP labeled ``morphology'', ``syntax'', ``syntax\_semantics'', and ``semantics'' by the authors of this benchmark. The evolution of performance split across these four categories is displayed on Supplementary Fig.~\ref{fig:olmo-2_rv_blimp}. The emergence of left-right asymmetry aligns slightly more closely with the performance on syntactic tests than with those pertaining to morphology or semantics. This suggests a particular salience of syntactic processing in driving the observed brain-LLM alignment. 

\subsection{Experiment 2: Further tests of formal and functional competence}

\begin{figure}[!h]
\centering
\includegraphics[width=\linewidth]{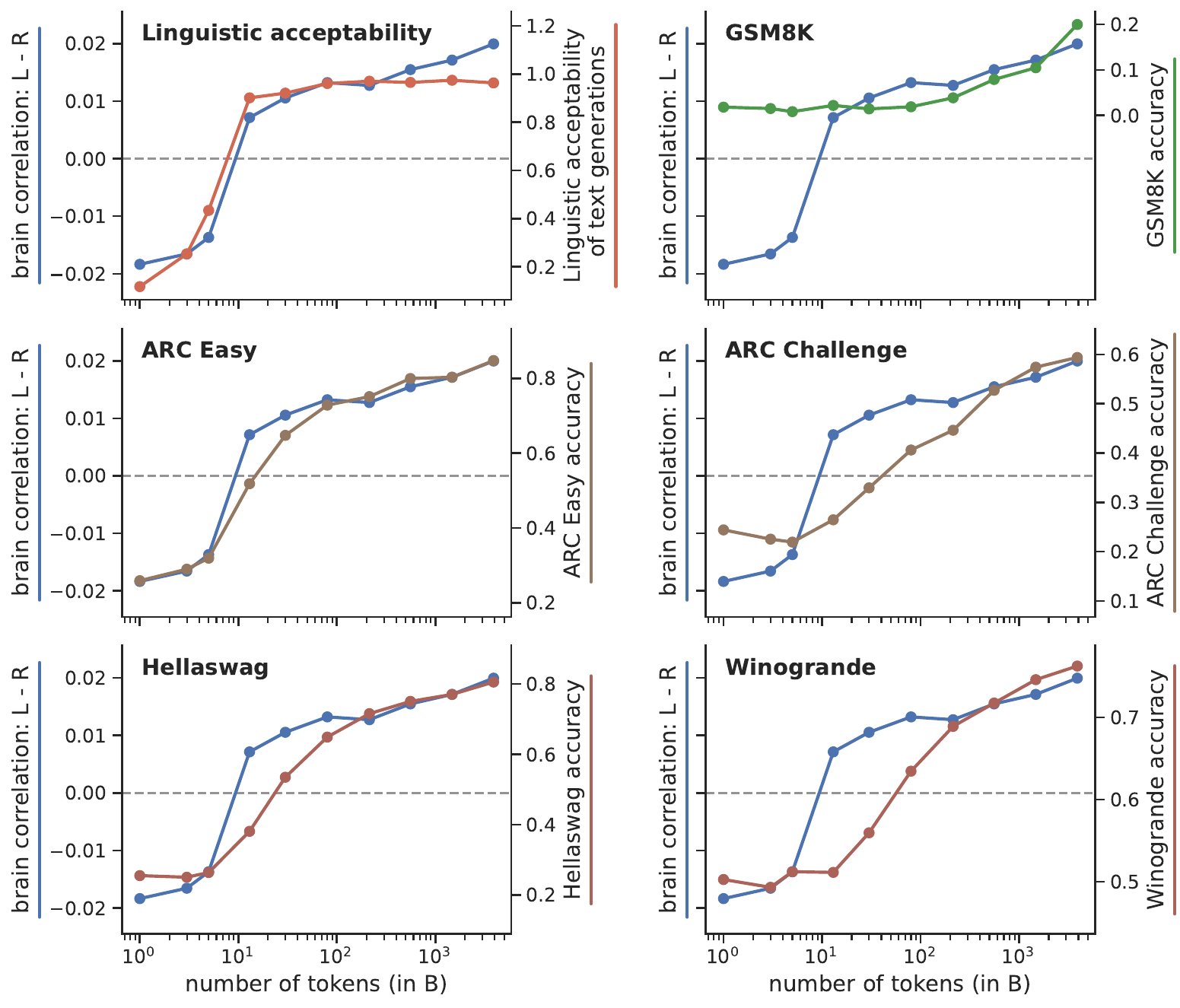} 
\caption{\textbf{Left-right hemispheric asymmetry aligns with the acquisition of formal linguistic competence, but not with high-level language comprehension nor mathematical reasoning capabilities}. Formal competence is assessed by automatically evaluating the linguistic acceptability of text generated at each training checkpoint. GSM8K consists of grade school math problems, while the ARC, Hellaswag, and WinoGrande benchmarks assess world knowledge and commonsense reasoning.}
\label{fig:olmo-2_rv_extra}
\end{figure}

Fig.~\ref{fig:olmo-2_rv_extra} shows the results of six additional tests: one (acceptability) assessing the formal linguistic competence of the model and the others assessing its functional competence (GSM8K, ARC Easy, ARC Challenge, Hellaswag, and Winogrande). Only the linguistic acceptability of texts generated by the model at various checkpoints exhibits a transition between 5B and 13B tokens that closely matches the left-right asymmetry. Examples of texts generated by the model at successive checkpoints, presented in Supplementary Materials, confirm that it is in this training range that the model starts to produce well-formed prose.

\subsection{Quantitative analysis of the alignment between trajectories}
\label{sec:methods_sigmoid}
The performance on the various benchmarks increases with the amount of tokens seen during training, as does the left-right asymmetry. On a $x$-log scale, this results in sigmoid shaped curves. To provide a quantitative comparison between all the different trajectories, for each curve displayed on Fig.~\ref{fig:olmo-2_rv_acc} and~\ref{fig:olmo-2_rv_extra}, we fit a sigmoid in order to locate the phase transition $x_0$ on the $x$-axis (log of the number of tokens) and its slope $\beta$. The fit is obtained by minimizing the mean square error between the target relevant curve and the following sigmoidal function: $y = y_{\min} + (y_{\max}-y_{\min})/(1 + \mbox{exp}({-\beta (x-x_0)})$, where $x$ is the logarithm of the number of tokens seen during training. Supplementary Fig.~\ref{fig:olmo-2_rv_sigmoid_fit} provides a full visualization of these fits.

\begin{figure}[!h]
\centering
\includegraphics[width=\linewidth]{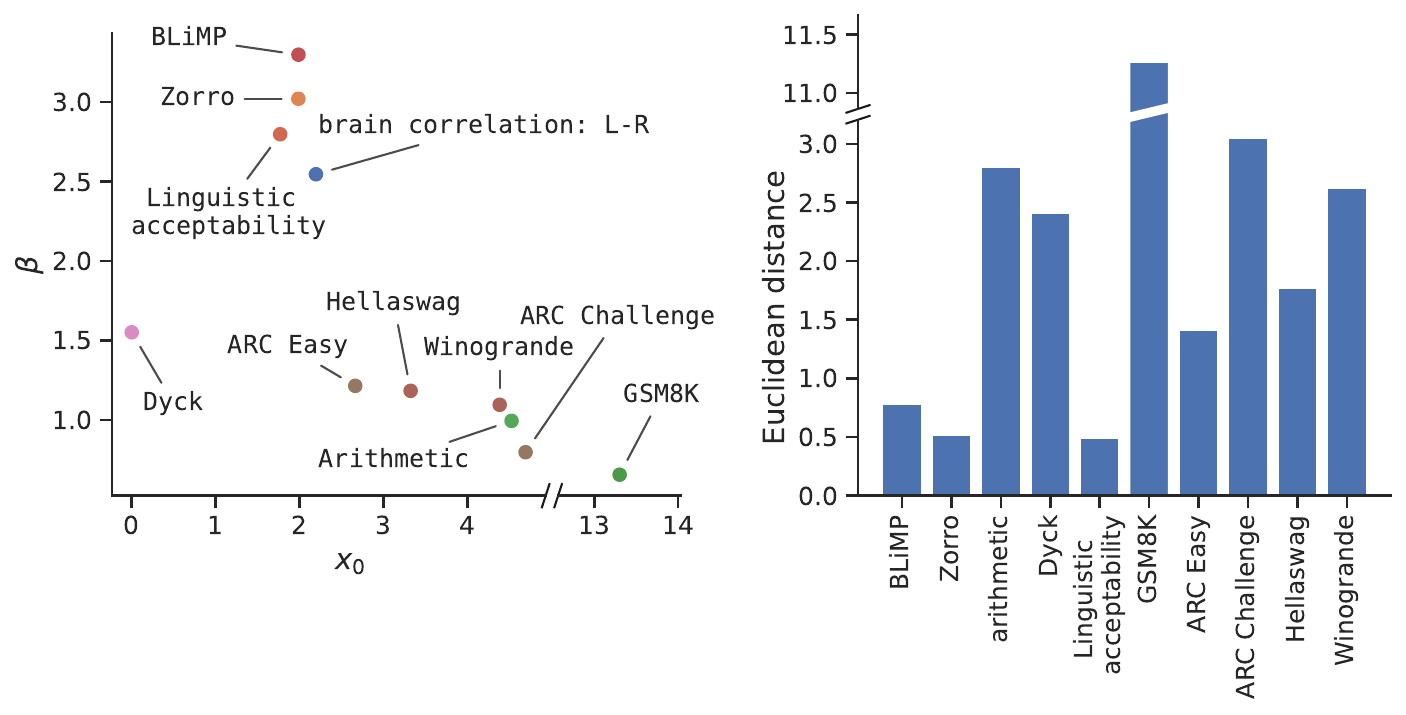}%
\caption{\textbf{Quantitative comparison of the evolution of the left-right hemispheric brain correlations and the various performance trajectories.} (Left) After fitting a sigmoid to the evolution of a given quantity, we plot the results on a $(x_0, \beta)$ plane, where $x_0$ is the location of the transition along the log(number of tokens) axis, and $\beta$ the slope of the change. (Right) Euclidean distance between the location on the $(x_0, \beta)$ plane of each benchmark and the left-right asymmetry.}
\label{fig:olmo-2_rv_sigmoid}
\end{figure}

The location $x_0$ of the transition and its slope $\beta$ can then be used to quantitatively compare all the different transitions. Panel (a) of Fig.~\ref{fig:olmo-2_rv_sigmoid} shows the location of each benchmark in the ($x_0, \beta$) space; Panel (b) shows the distance of each benchmark to the parameters of the brain asymmetry transition. This quantitatively confirms that the left-right asymmetry aligns well with the acquisition of formal linguistic competence, but not with high-level language comprehension or other competences such as arithmetic ability.

\subsection{Experiment 3: Replications with two Pythia models}
\label{sec:pythiaresults}
\begin{figure}
\centering
\hfill
\includegraphics[height=.34\linewidth, valign=t]{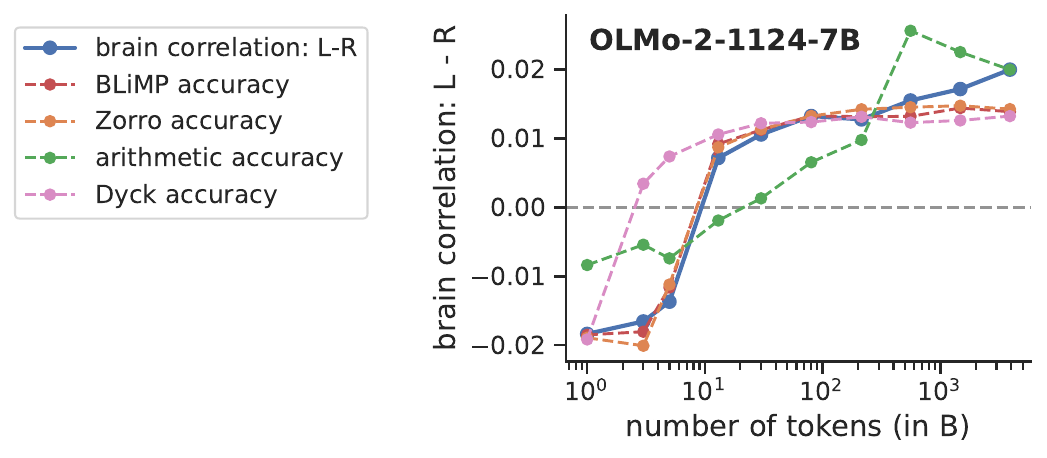}\\[12pt]
\includegraphics[height=.34\linewidth]{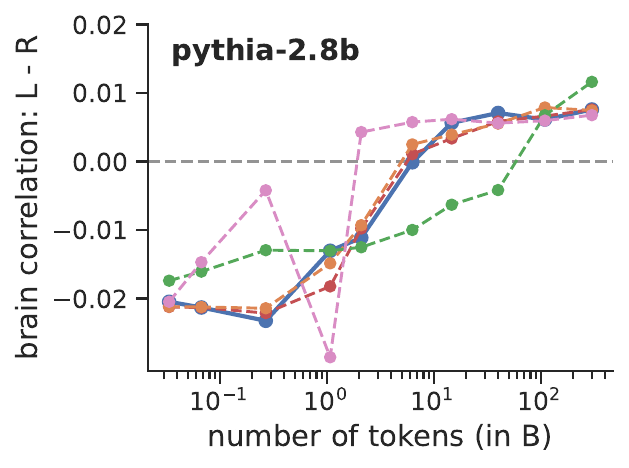}%
\hfill
\includegraphics[height=.34\linewidth]{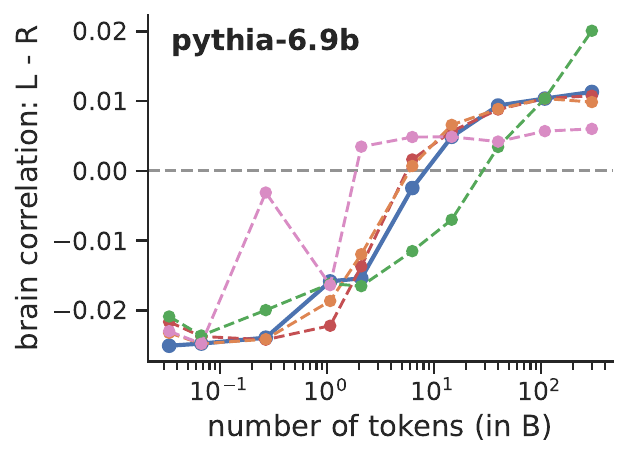}%
\caption{\textbf{Generalization to other language models.} Results from minimal-pair benchmarks on OLMo-2-1124-7B (top panel) extend to Pythia-2.8b (bottom left panel) \& 6.9b models (bottom right panel). Note the OLMo-2 panel reproduces data shown in Fig.~\ref{fig:olmo-2_rv_acc} but with all curves superimposed (Fig.~\ref{fig:pythia_rv_acc} provides figures split by tasks for Pythia models).}
\label{fig:3models_rv_acc}
\end{figure}

To check that the results are not specific to the OLMo-2 7B model, we replicate Experiment 1 with two models from the Pythia family, which also provides checkpoints during training. Here again, as shown on Fig.~\ref{fig:3models_rv_acc}, the left-right asymmetry aligns remarkably well with the acquisition of formal linguistic competence by the model (assessed by BliMP and Zorro benchmarks), but not with the performance on the functional benchmarks (arithmetic and Dyck).

\subsection{Experiment 4: Replication in two other languages (French and Chinese)}
Are the previous results specific to English or do they replicate in other languages? We investigate this question using the French and Chinese fMRI datasets available from the \textit{Le Petit Prince} project. As we did for English, we evaluate the main model, OLMo-2 7B, on French and Chinese, both on formal and functional benchmarks.

\begin{figure}
\textbf{\large A}\hspace{-0.1cm}
\includegraphics[width=\linewidth, valign=t]{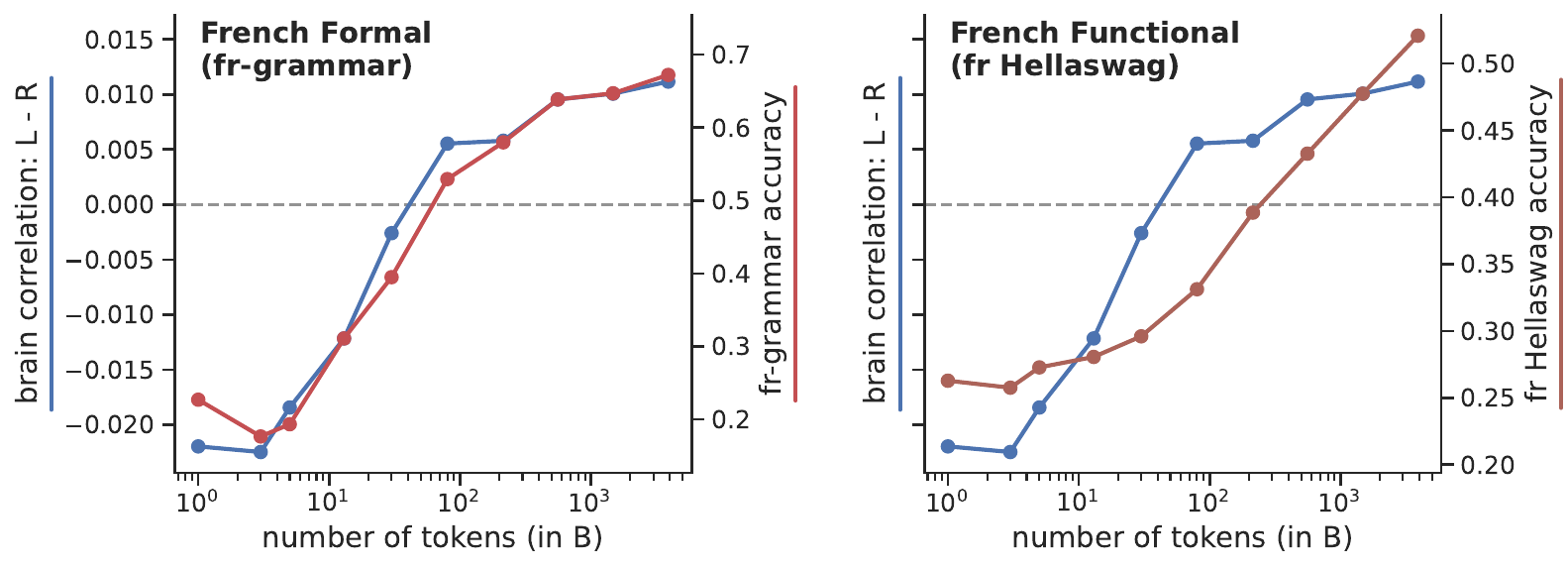}\\[15pt]%
\textbf{\large B}\hspace{-0.1cm}
\includegraphics[width=\linewidth, valign=t]{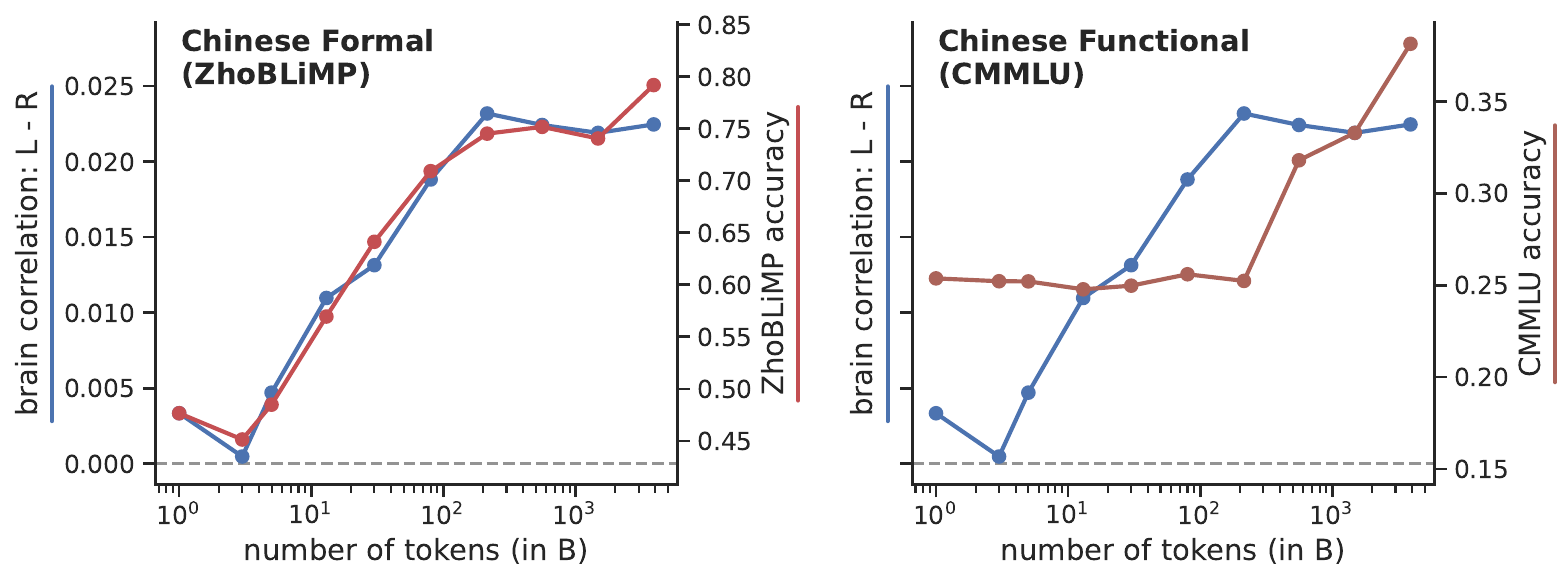}%
\caption{\textbf{The evolution of the left-right asymmetry in brain predictivity in French (A) and Chinese (B) follows OLMo-2-1124-7B's acquisition of formal competence in each language}. The blue lines represent the evolution of the left-right difference in brain correlations computed on the French (top row, A) and the Chinese (bottom row, B) average subject.
In the left panels, the red curves track performance on formal linguistic competence in these languages, evaluated with the fr-grammar benchmark (from the FrenchBench evaluation benchmark, \citealp{faysse2025croissantllm}) for French, and with ZhoBLiMP \citep{liu2024systematic} for Chinese. In the right panels, the brown curves show the evolution of functional competences of the language model, assessed with French Hellaswag \citep{faysse2025croissantllm} and CMMLU\citep{li2024cmmlu}.}
\label{fig:olmo-2_rv_fr_cn}
\end{figure}

Results are displayed on Fig.~\ref{fig:olmo-2_rv_fr_cn}. Similar to the findings for English, the evolution of formal competence follows more closely the left-right asymmetry than does the performance on benchmarks assessing higher language understanding (see also Supplementary Fig.~\ref{fig:olmo-2_rv_sigmoid_fr_cn} for a quantitative assessment of the alignments).

As expected given that OLMo-2 7B was primarily trained on English, the development of its formal competence in French or Chinese is delayed and progresses more slowly compared to English, and so does the left-right hemispheric asymmetry in brain correlations. Supplementary Fig.~\ref{fig:olmo-2_rv_sigmoid_fr_cn}, comparing the parameters of fitted sigmoids of all the relevant quantities, confirms this.

\subsection{Experiment 5: Right-left hemispheric asymmetry in the cerebellum} 

\begin{figure}
\centering
\includegraphics[width=.97\linewidth, valign=t]{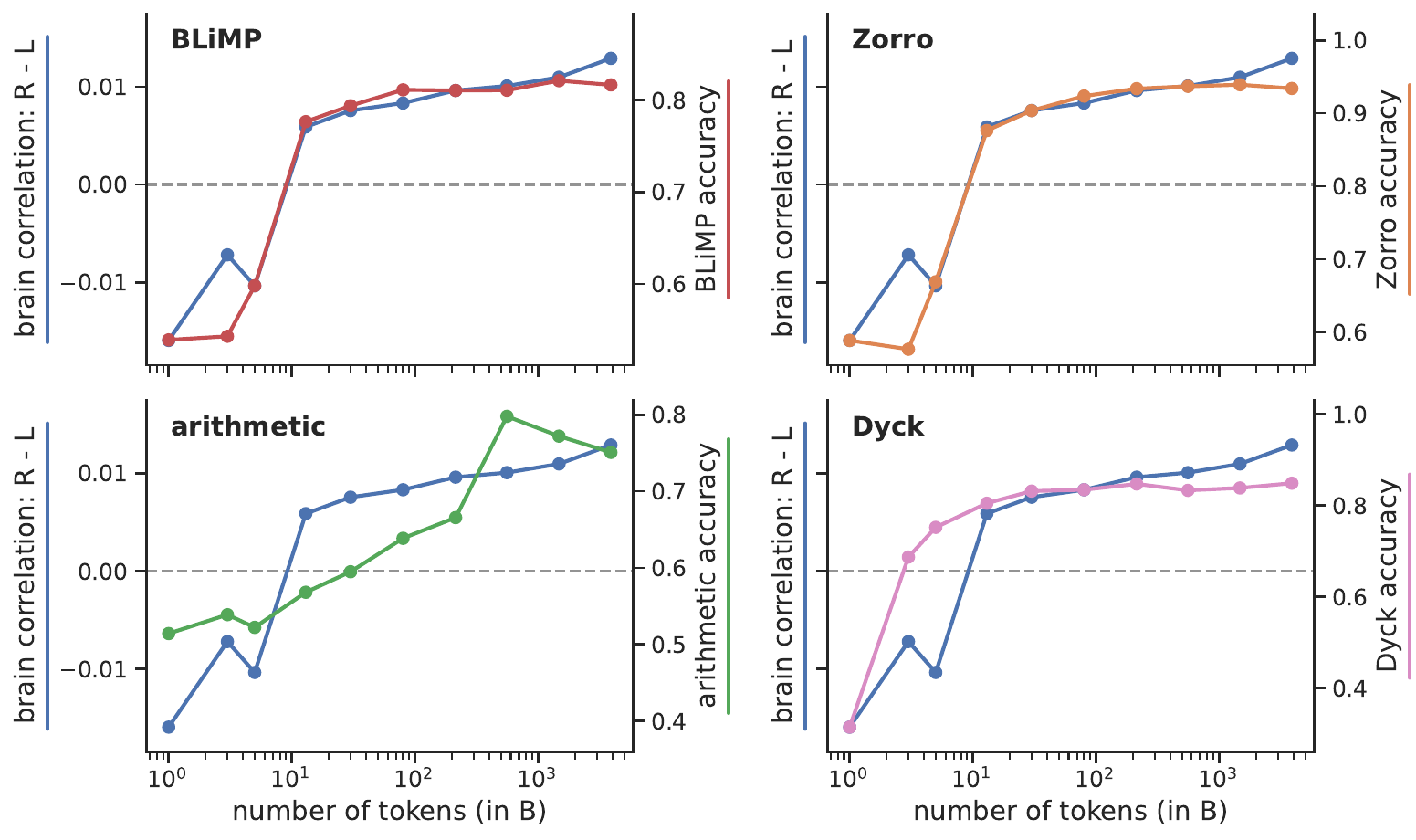}%
\caption{\textbf{Evolution of the right-left asymmetry in the cerebellum. Phase transition during training: minimal-pair benchmarks.} Same as Fig.~\ref{fig:olmo-2_rv_acc}, but for the cerebellum. Note that the $y$-axis represents here the \emph{right-left} difference in cerebellar hemispheres, contrary to the original figure showing the left-right difference. See Supplementary Fig.~\ref{fig:cerebellum_mask} for a visualization of the left and right cerebellar symmetric masks.}
\label{fig:olmo-2_cerebellum}
\end{figure}

Fig.~\ref{fig:olmo-2_cerebellum} shows that the case of cerebellum nicely mirrors the one found previously: a right-left asymmetry in brain predictivity emerges during the training of the LLM, and this phase transition remarkably aligns with the acquisition of formal linguistic competence by the model.

\subsection{Robustness to mask choice}

In experiments 1 to 4, the left-right asymmetry is computed from a mask based on the 25\% voxels with the highest inter-subject correlations (ISC). In order to check whether the results depend on this specific choice, we computed the left-right hemisphere asymmetry using different approaches: a whole brain mask, a symmetrized mask based on the 25\% voxels in both hemisphere (also normalizing by average ISC over each hemisphere), and a mask constructed from Fedorenko's lab language regions of interest available at \url{https://www.evlab.mit.edu/resources-all/download-parcels}, which are updates of the original parcels described in \citep{fedorenko2010new}. These masks are displayed on Supplementary Fig.~\ref{fig:olmo-2_brain_corr_l_r}, which also provides more details on their computations. The figure also presents for each mask the raw brain-predictivity scores obtained in the left and right hemispheres separately. 

Supplementary Fig.~\ref{fig:olmo-2_l_r_asym_comparison} reports the left-right asymmetries computed with the various approaches. Overall, one can see that the transition of the left-right asymmetry is robust across the various methods of computation (the only exception is the case of the whole brain mask with the Chinese dataset (right top panel), which we further investigate in Supplementary Fig.~\ref{fig:olmo-2_l_r_asym_urv}).

\section{Discussion}

It is now well established that large language models can, to some extent, predict brain activation during language processing. Why this is the case remains a largely open question and the current work is a step towards answering it. It stems from the observation that during training, LLMs progressively predict relatively better the activations in the left hemisphere compared to the right hemisphere. What information is discovered by the LLM that explains this asymmetry?

In order to understand the origin of the emergence of the left-right asymmetry in the correlation between brain activity and model predictions with training, we ran a number of benchmarks on LLMs (OLMo-2 7B, Pythia 2.8b, and Pythia 6.9b) at different training checkpoints. First, we reproduce and extend  our previous finding that the left-right asymmetry emerges with training \citep{bonnasse-gahot_fmri_2024} to new models and with a more fine-grained resolution of training steps. Second, we show that the left-right asymmetry emergence co-occurs with the emergence of formal linguistic abilities in LLMs, attested either by their ability to assign a higher probability to an acceptable sentence than to a grammatically unacceptable one within a minimal contrasting pair (BLiMP and Zorro benchmarks on Fig.~\ref{fig:olmo-2_rv_acc}), or their capacity to produce well-formed text (Fig.~\ref{fig:olmo-2_rv_extra}). Furthermore, the trajectory of the left-right asymmetry with training did not align with arithmetic or formal language (Dyck) tasks (Fig.~\ref{fig:olmo-2_rv_acc}), nor with tasks involving world knowledge and reasoning (ARC, Hellaswag, WinoGrande, and GSM8K; see Fig.~\ref{fig:olmo-2_rv_extra}). Finally, we replicate this result in two other languages, French and Chinese.  

In a recent study, \citet{alkhamissi2025language} compared the developmental trajectories of brain correlations, formal linguistic competence and functional competence \citep{mahowald2024dissociating} and showed three successive phase transitions: brain scores raise first, followed by formal competence, and only later by functional competence. Here, we also find that functional competence is acquired later during training compared to formal competence, but we find that the left-right asymmetry strikingly aligns with the trajectory of formal performance (see Figs.~\ref{fig:olmo-2_rv_acc},~\ref{fig:olmo-2_rv_extra}, ~\ref{fig:3models_rv_acc}, and~\ref{fig:olmo-2_rv_fr_cn}), contrary to the absolute brain-predictivity score which start to increase before formal competence \citep[see][Fig.~4]{alkhamissi2025language}.

Among all our tasks, one was especially easy to acquire: the Dyck languages based on nested parentheses. One possibility is that this is due to in-context learning: \citet{olsson2022context} proposed that some attention heads (``induction heads'') enable a model to recognize and complete patterns based on previous occurrences in a prompt. They reported that transformer language models undergo a ``phase change'' early in training, during which induction heads form and simultaneously in-context learning improves dramatically. This mechanism could be at play for the Dyck languages in our experiment. Another possibility could be due to low-level reasons such as bigrams violations in the ungrammatical sequences of parentheses (e.g. \{ ) in Dyck-3). In any case, the underlying phenomena is acquired early by the LLM, well before the left-right transition.

We have shown that, as training progresses, using the internal representations of an LLM to predict brain activity leads to a sudden increase in asymmetry between left and right hemispheres, coinciding with the LLM's acquisition of formal linguistic competence, at around 10B tokens. After this main phase transition, the functional competence of the LLM keeps improving (see scores on ARC, Hellaswag, WinoGrande, and GSM8K benchmarks on Fig.~\ref{fig:olmo-2_rv_extra}). One can see that the left-right asymmetry also continues to grow slightly (see Fig.~\ref{fig:olmo-2_rv_acc}), following the increase in performance on higher level language understanding tasks (reasoning and world knowledge). More work is needed to understand the development of more sophisticated linguistic capabilities and how it translates into the alignment between artificial and biological neural processing.

One point of caution is in order. One should not jump to the conclusion that brain correlations are only driven by syntactic knowledge. \citet{kauf_lexical-semantic_2024}, manipulating sentences by altering word order, removing words, or changing semantic content, observed that brain correlations were more affected by changing semantic content. This led them to claim that ``lexical-semantic content of the sentence (largely carried by content words) rather than the sentence's syntactic form (conveyed via word order or function words) is primarily responsible for the ANN-to-brain similarity''. It would be interesting to check how these manipulations impact the left-right asymmetry.

Although we focused in this paper on a global property, the left-right hemispheric asymmetry, the relationship between correlation between brain activity and model predictions and the linguistic performance of models at different training stages should eventually be evaluated more finely at the level of brain regions. This type of approach has been applied very recently to visual processing by \citet{raugel2025disentangling}, who observed that brain-predictivity scores in various regions have different trajectories as a function of the amount of training. More precisely, they reported that the model they study, Dino v3, first aligns with the early representations of the sensory cortices, and needs more training data to align with higher-level regions. Future work will address whether similar links between brain regions and LLMs at different training steps exist for language.

\section{Data and Code Availability Statements}
All fMRI data come from the publicly available fMRI corpus \textit{Le Petit Prince} \citep{li2022petit} (\url{https://openneuro.org/datasets/ds003643/versions/2.0.5}). The code is available on GitHub: \url{https://github.com/l-bg/llm_training_brain_asym}. 
It relies on Python 3.10 and the following libraries: 
\texttt{transformers v4.56.0} \citep{wolf2020transformers},
\texttt{scikit\_learn v1.6.1} \citep{pedregosa2011scikit},
\texttt{nilearn v0.11.1},
\texttt{Pytorch v2.7.1} \citep{paszke2019pytorch},
\texttt{nltk v3.9.1} \citep{bird2009natural},
\texttt{matplotlib v3.10.3} \citep{hunter2007matplotlib},
\texttt{seaborn v0.13.2} \citep{waskom2021seaborn},
\texttt{numpy v2.0.2} \citep{van2011numpy},
\texttt{pandas v2.2.3} \citep{mckinney2010data},
\texttt{scipy v1.15.2} \citep{scipy}.
All pretrained models were downloaded from Hugging Face through the \texttt{transformers} interface. To assess the performance of OLMo-2 7B on GSM8K, ARC, Hellaswag, WinoGrande, FrenchBench, and CMMLU, we rely on EleutherAI's evaluation tools \texttt{lm\_eval v0.4.9}~\citep{eval-harness}, available at \url{https://github.com/EleutherAI/lm-evaluation-harness}.

\section{Acknowledgements}
We would like to thank the two anonymous reviewers for their constructive comments that helped to improve the paper.

\clearpage
\bibliographystyle{apalike}
\bibliography{refs.bib}

\clearpage
\appendix

\renewcommand\thefigure{S\arabic{figure}}  
\setcounter{figure}{0}
\renewcommand{\theHfigure}{\thefigure}

\section*{Supplementary Figures}
\label{sec:supp_fig}

\begin{figure}[h]
\begin{tabular}{ccc}
\includegraphics[width=0.3\linewidth]{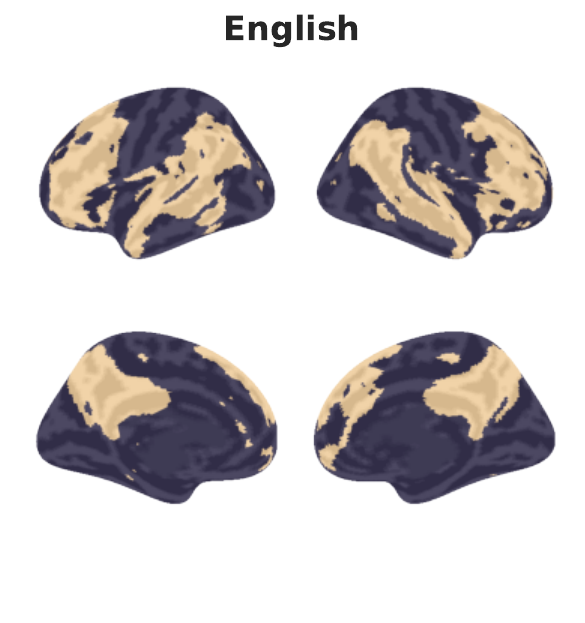} &
\includegraphics[width=0.3\linewidth]{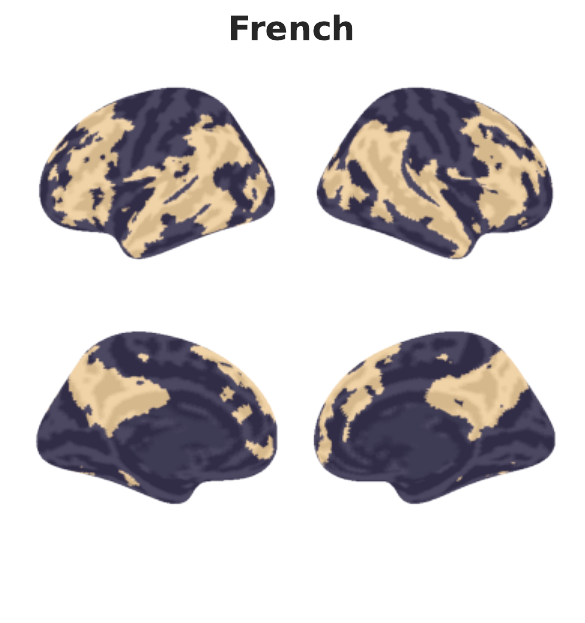} &
\includegraphics[width=0.3\linewidth]{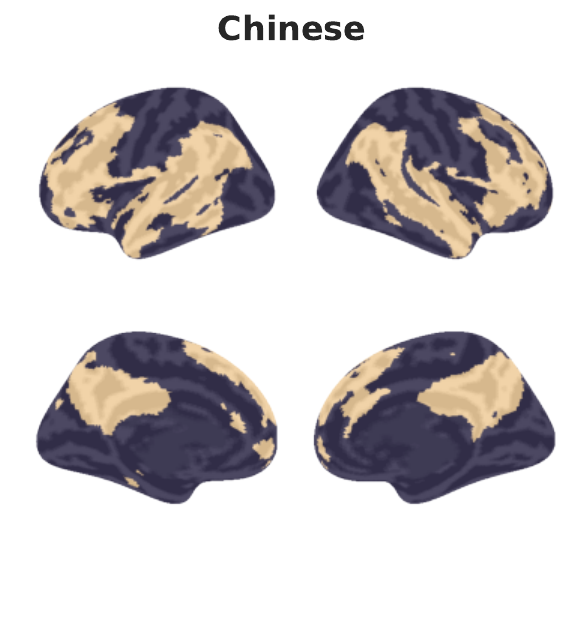} \\
\end{tabular}\\[12pt]
\caption{\textbf{Cortical surface renderings of the masks used to extract brain predictivity scores used in all main figures of the paper}. The masks encompass the voxels having the 25\% strongest inter-subject-correlations within a whole brain mask (the figures here do not display the cerebellum, which also included some voxels with high ISC). Note that these masks, obtained from three independent datasets (different machines, different participants, different languages), are similar, essentially symmetrical and encompass the core temporo-frontal language network.}
\label{fig:masks}
\end{figure}

\begin{figure}[hb]
\centering
\includegraphics[width=\linewidth]{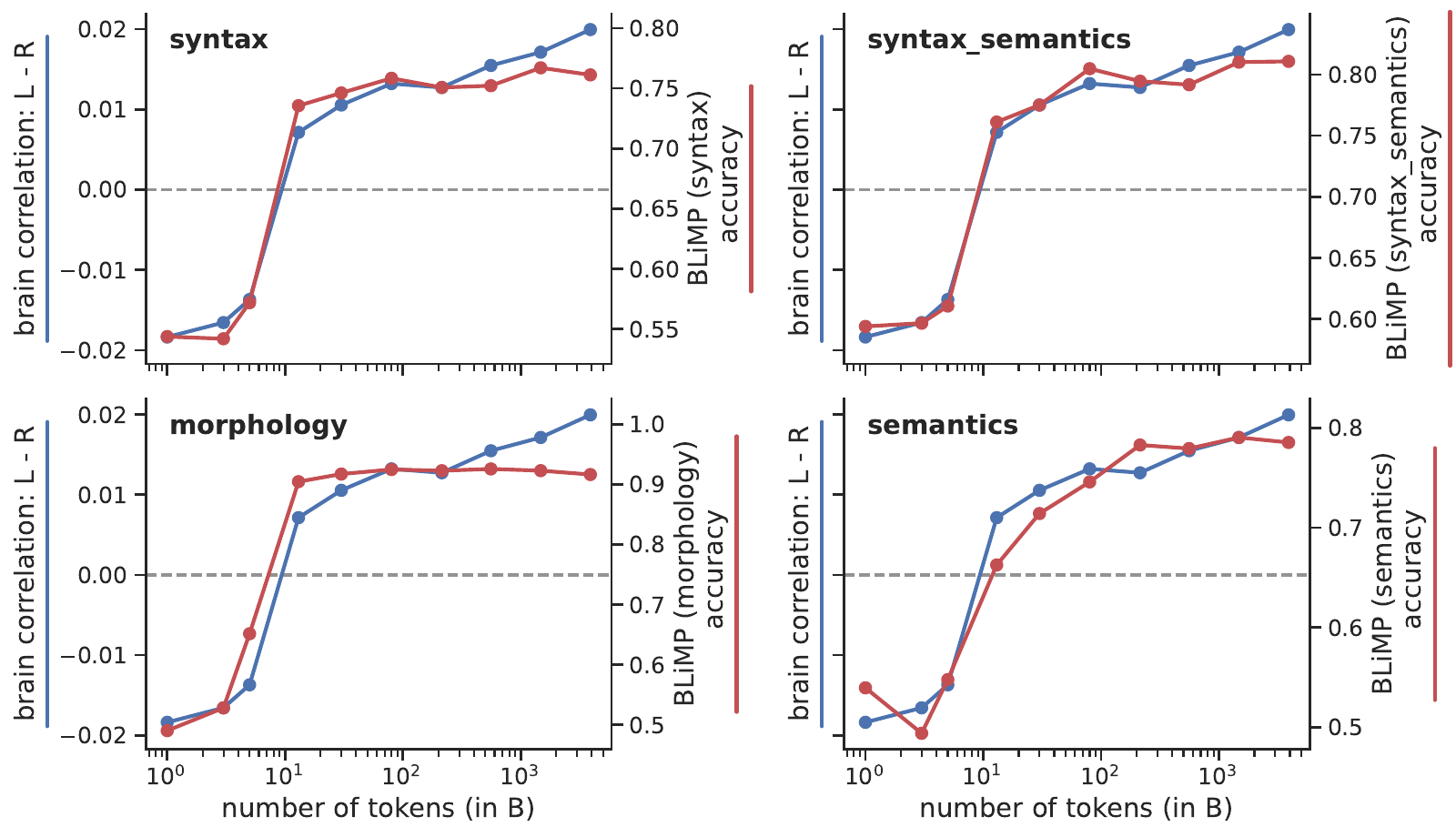} 
\caption{\textbf{Results split by BLiMP subtasks}. BLiMP categorizes minimal pairs into four different fields (``syntax'', ``syntax\_semantic'', ``morphology'', ``semantic''). For each field, a panel displays the evolution of performance on the corresponding minimal pairs, along with the left-right brain predictivity asymmetry (as in Fig.~\ref{fig:olmo-2_rv_acc}).}
\label{fig:olmo-2_rv_blimp}
\end{figure}

\begin{figure}
	\centering
	\includegraphics[width=\linewidth]{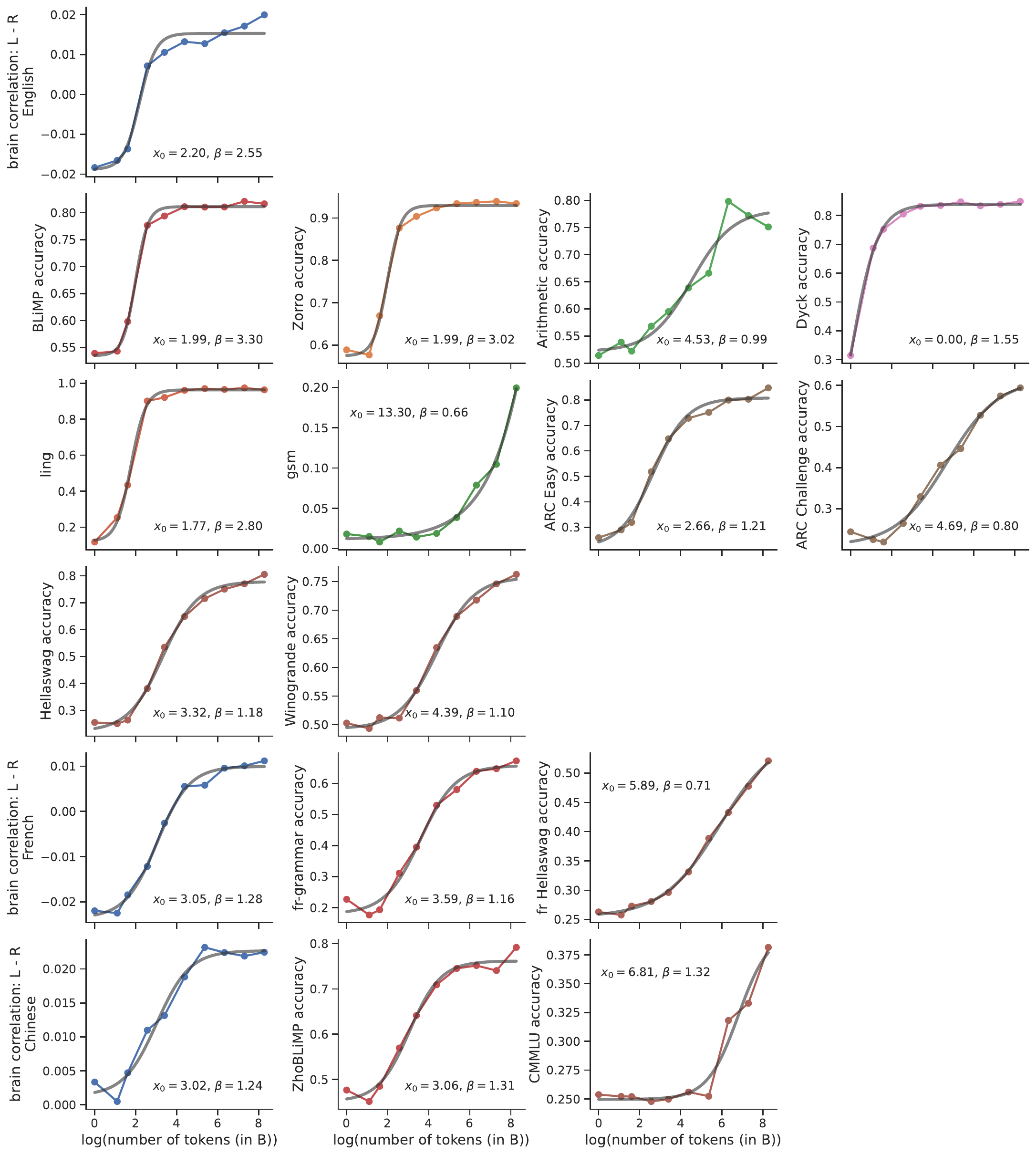} 
	\caption{\textbf{Sigmoid fits to the evolution of various quantities during the training of the OLMo-2 7B language model}. The colored lines correspond to empirical data, while the gray lines are the fitted sigmoidal curves. The $x_0$ and $\beta$ (slope) parameters estimated from these fits were used to construct Fig.~\ref{fig:olmo-2_rv_sigmoid}.}
	\label{fig:olmo-2_rv_sigmoid_fit}
\end{figure}

\begin{figure}[h]
\centering
\textbf{Pythia-2.8b}\\[5pt]
\includegraphics[width=\linewidth]{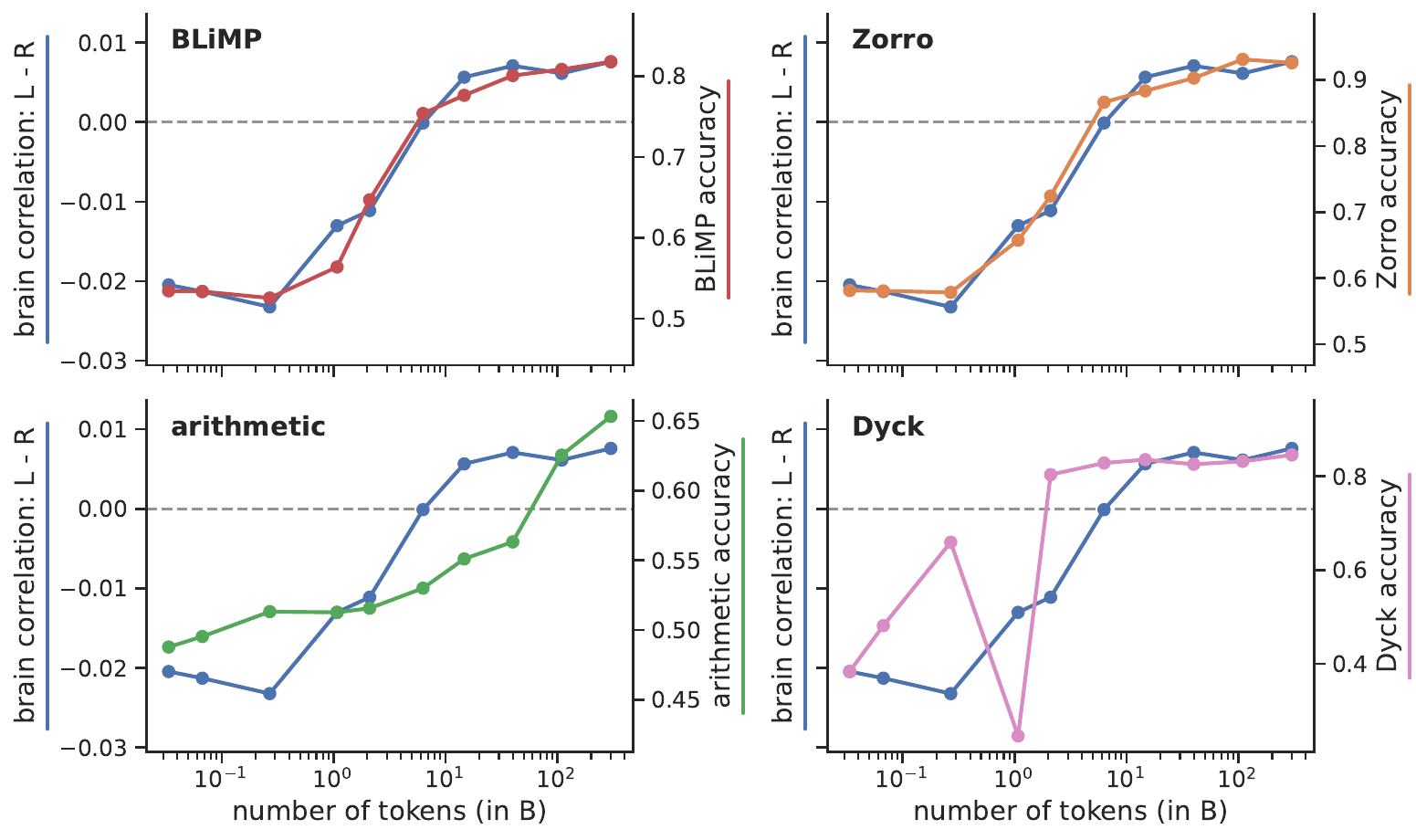}\\[10pt]
\textbf{Pythia-6.9b}\\[5pt]
\includegraphics[width=\linewidth]{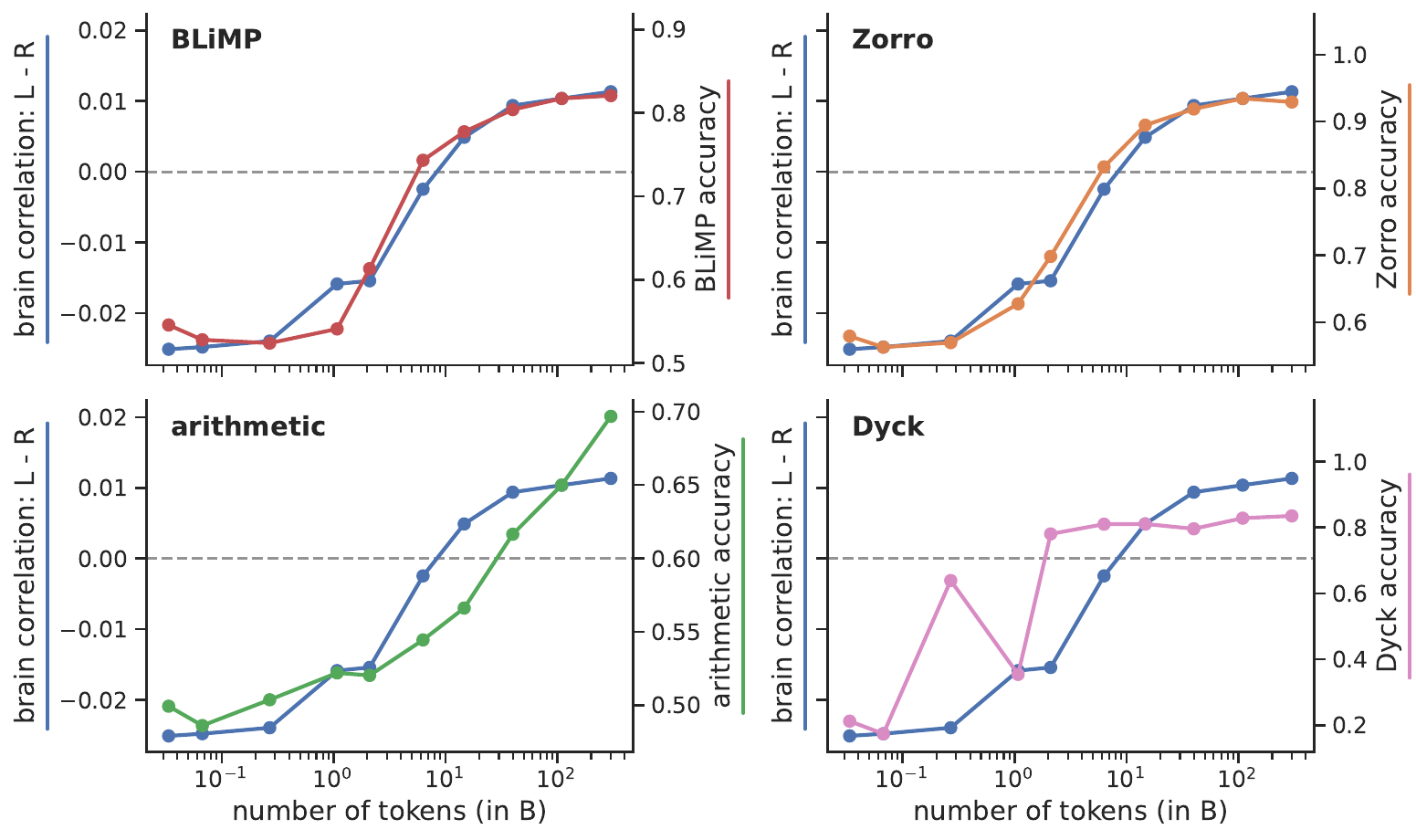} 
\caption{\textbf{Generalization to other language models.} Same as Fig.~\ref{fig:olmo-2_rv_acc} but for the Pythia-2.8b (top panel) and Pythia-6.9b (bottom panel) models. In each case, the trajectory of left-right brain asymmetry aligns well with the evolution of the performance on linguistic minimal-pair benchmarks, but not on the non-linguistic ones.}
\label{fig:pythia_rv_acc}
\end{figure}

\begin{figure}
\centering
\textbf{\large A}\hspace{-0.1cm}
\includegraphics[width=.77\linewidth,valign=t]{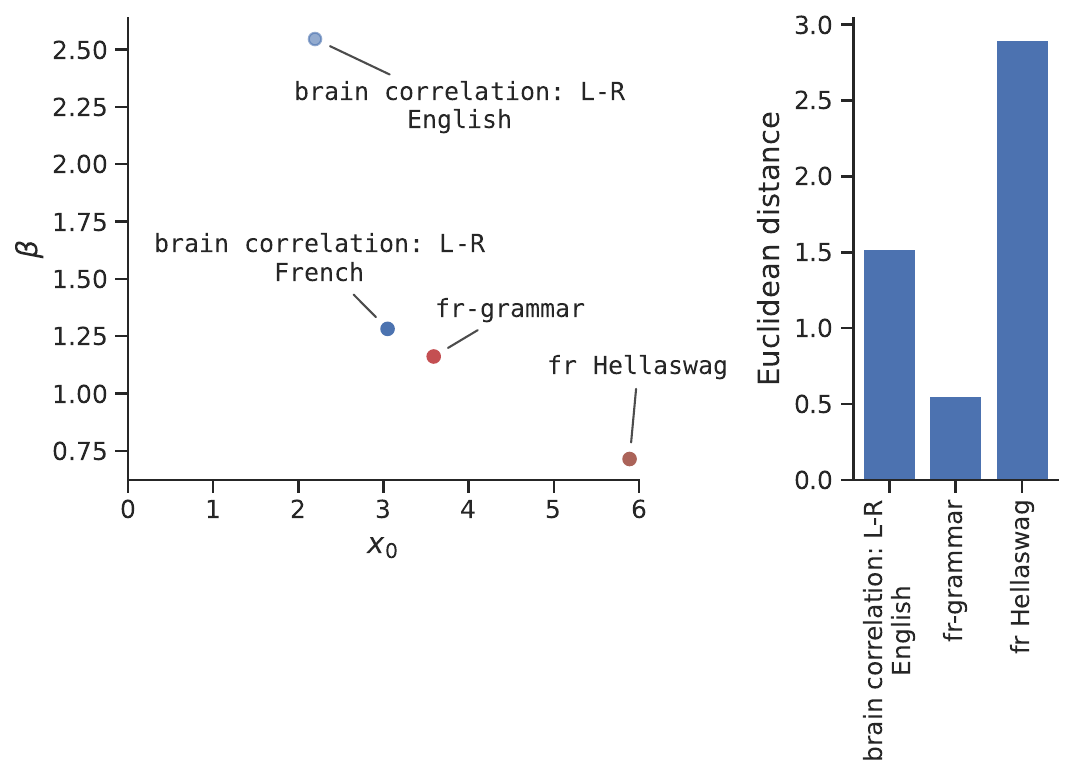}\\[15pt]%
\textbf{\large B}\hspace{-0.1cm}
\includegraphics[width=.77\linewidth,valign=t]{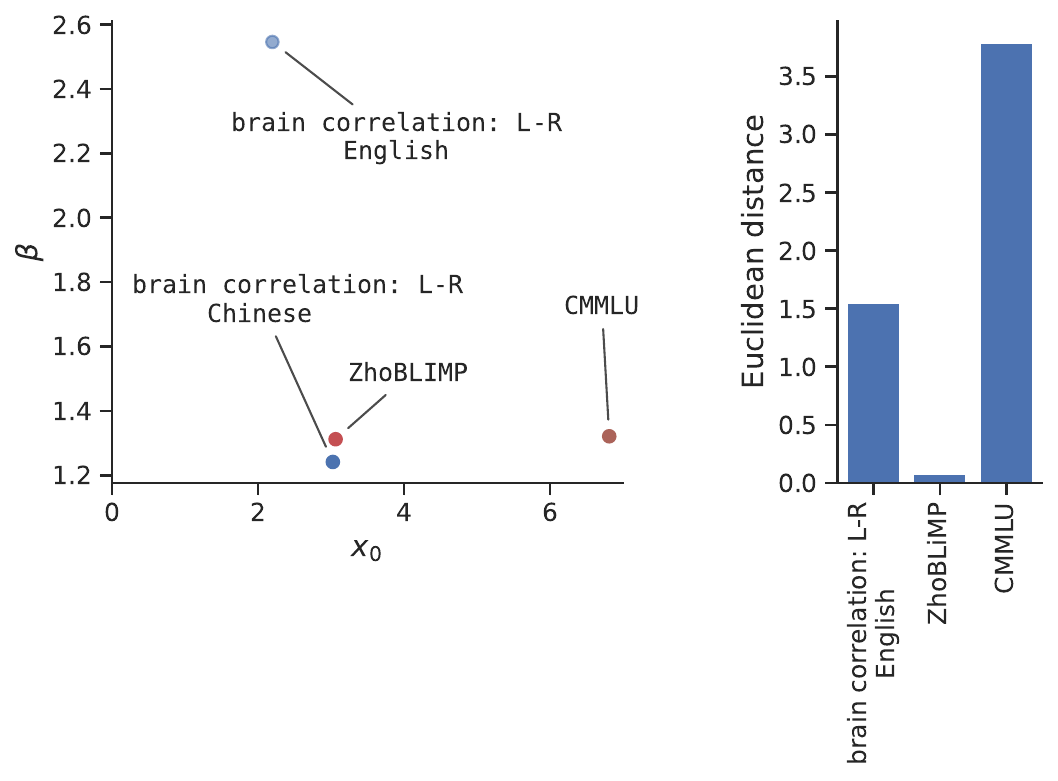}%
\caption{\textbf{Quantitative comparison of the evolution of the left-right hemispheric brain correlation in French (A) and Chinese (B) participants and the various performance trajectories.} Similar analysis as in Fig.~\ref{fig:olmo-2_rv_sigmoid} but for the French and Chinese data. Left-right asymmetry in English participants is also provided as a comparison. The right panel shows the Euclidean
distance between the location on the $(x_0, \beta)$ plane of each point on the left panel and the left-right asymmetry in French or Chinese participants. See Fig.~\ref{fig:olmo-2_rv_sigmoid_fit}, bottom rows, for a full visualization of the sigmoid fits.}
\label{fig:olmo-2_rv_sigmoid_fr_cn}
\end{figure}

\clearpage

\begin{figure}
\centering
\includegraphics[width=.75\linewidth, valign=t]{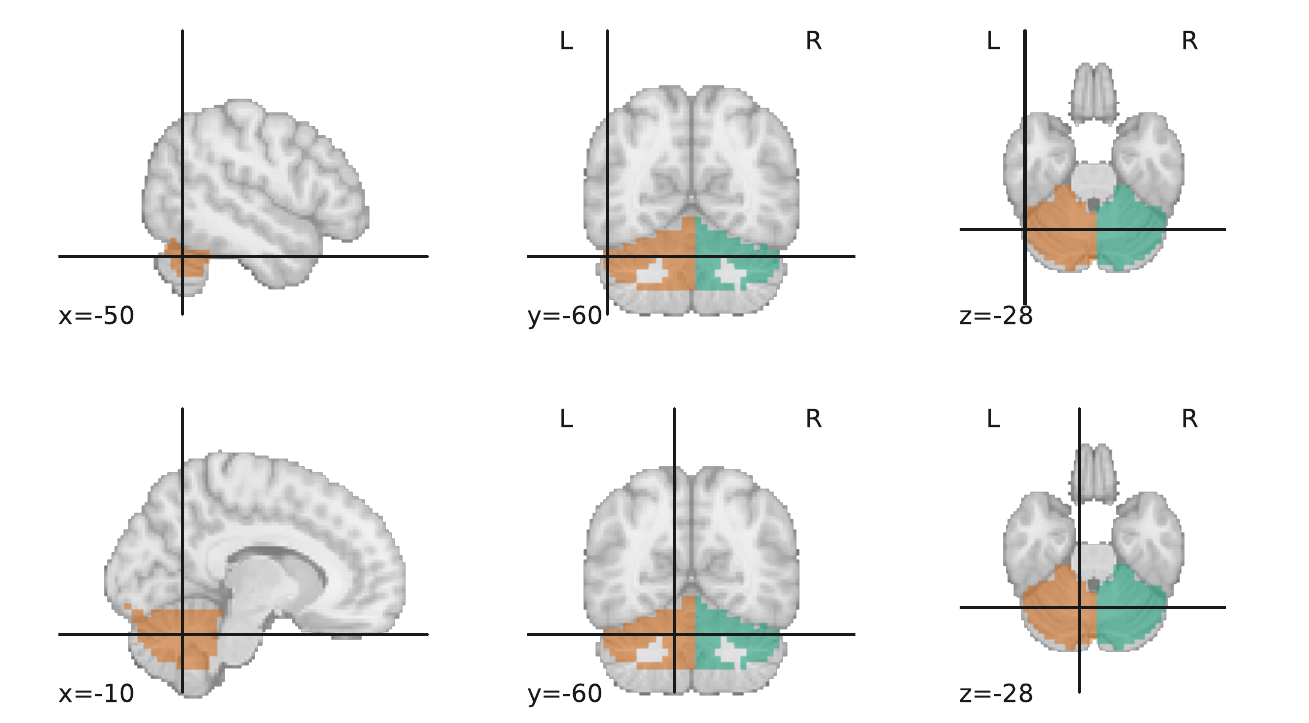}\\[16pt]
\caption{\textbf{Left and Right Cerebellum masks.} These masks were obtained from the symmetric BASC atlas scaled at~12 \citep{bellec2010multi}.} 
\label{fig:cerebellum_mask}
\end{figure}

\clearpage

\begin{figure}[!ht]
\centering
\includegraphics[width=\linewidth, valign=t]{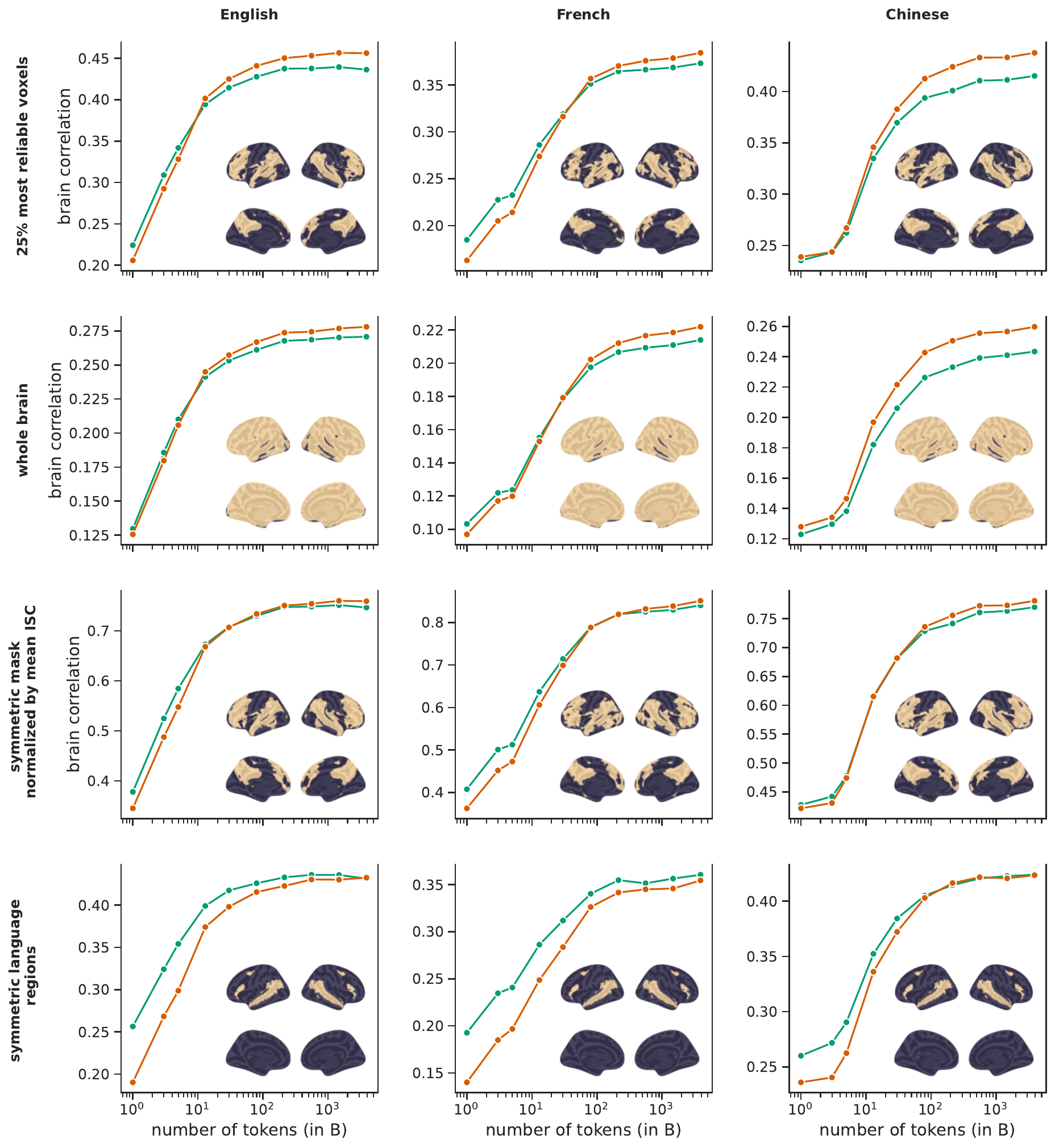}%
\caption{\textbf{Hemisphere-specific brain predictivity values (left and right separately), for various masks.}
Each column corresponds to a different language (computed from its corresponding text and fMRI average subject). 
(First row) Main mask used in the study, made of the 25\% voxels with the highest inter-subject correlations, also presented in Fig.~\ref{fig:masks}.
(Second row) Whole brain.
(Third row) Symmetric mask computed by taking the 25\% voxels with the greatest inter-subject correlations for each hemisphere separately, then combining them and symmetrizing the mask. The mean brain correlation for each hemisphere is then normalized by the mean ISC in its corresponding hemisphere. 
(Fourth row)
Symmetric mask of language areas, from the parcels provided by Fedorenko lab, which are updates of the original parcels described in \citep{fedorenko2010new}.
As for Fig.~\ref{fig:masks}, the masks are presented on the surface of the cortex only, but for the first three cases, there are also voxels in other brain regions, in particular in the cerebellum.
}
\label{fig:olmo-2_brain_corr_l_r}
\end{figure}

\clearpage

\begin{figure}
\centering
\includegraphics[width=\linewidth, valign=t]{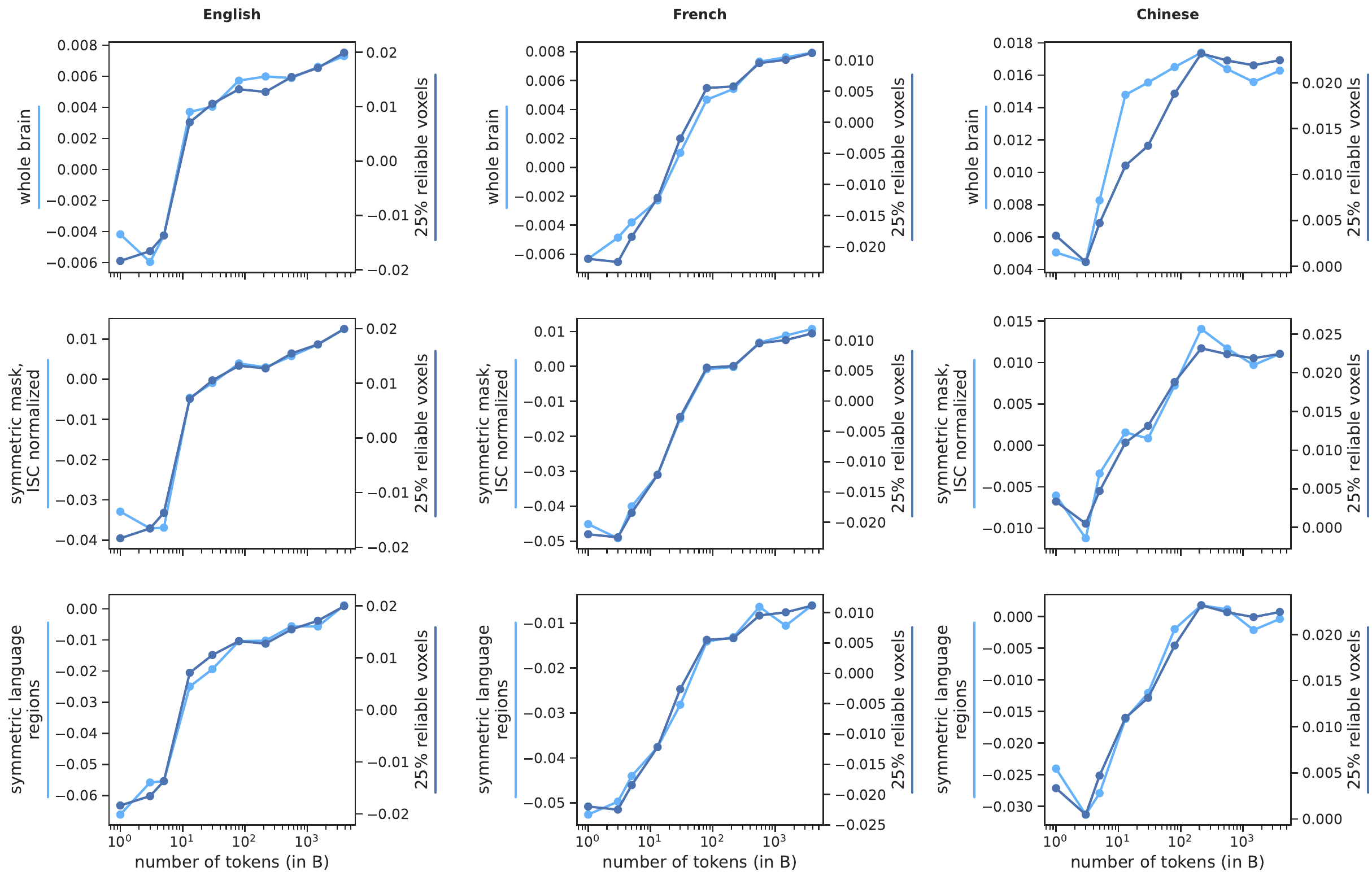}\\
\caption{\textbf{Robustness to mask choice: comparison of various methods to compute left-right asymmetry}.
The lighter blue curves, associated with the left $y$-axis, are variants of the left-right asymmetry computed from the masks displayed in Row 2 to Row 4 of Fig.~\ref{fig:olmo-2_brain_corr_l_r}.
The dark blue dotted curves represents the left-right asymmetry used in the paper, computed on the 25\% most reliable voxels for each language. Across the methods (whole brain, symmetric mask + ISC normalization, language regions), the left-right asymmetry in the correlation between brain activity and model prediction has a robust shape.
The only exception concerns the Chinese dataset for the whole-brain mask: the left-right asymmetry emerges a bit earlier than with the other variants. Focusing on `unreliable' voxels indeed shows a spike at this location during training, as shown in Fig.~\ref{fig:olmo-2_l_r_asym_urv}.
}
\label{fig:olmo-2_l_r_asym_comparison}
\end{figure}

\begin{figure}
\centering
\includegraphics[width=\linewidth, valign=t]{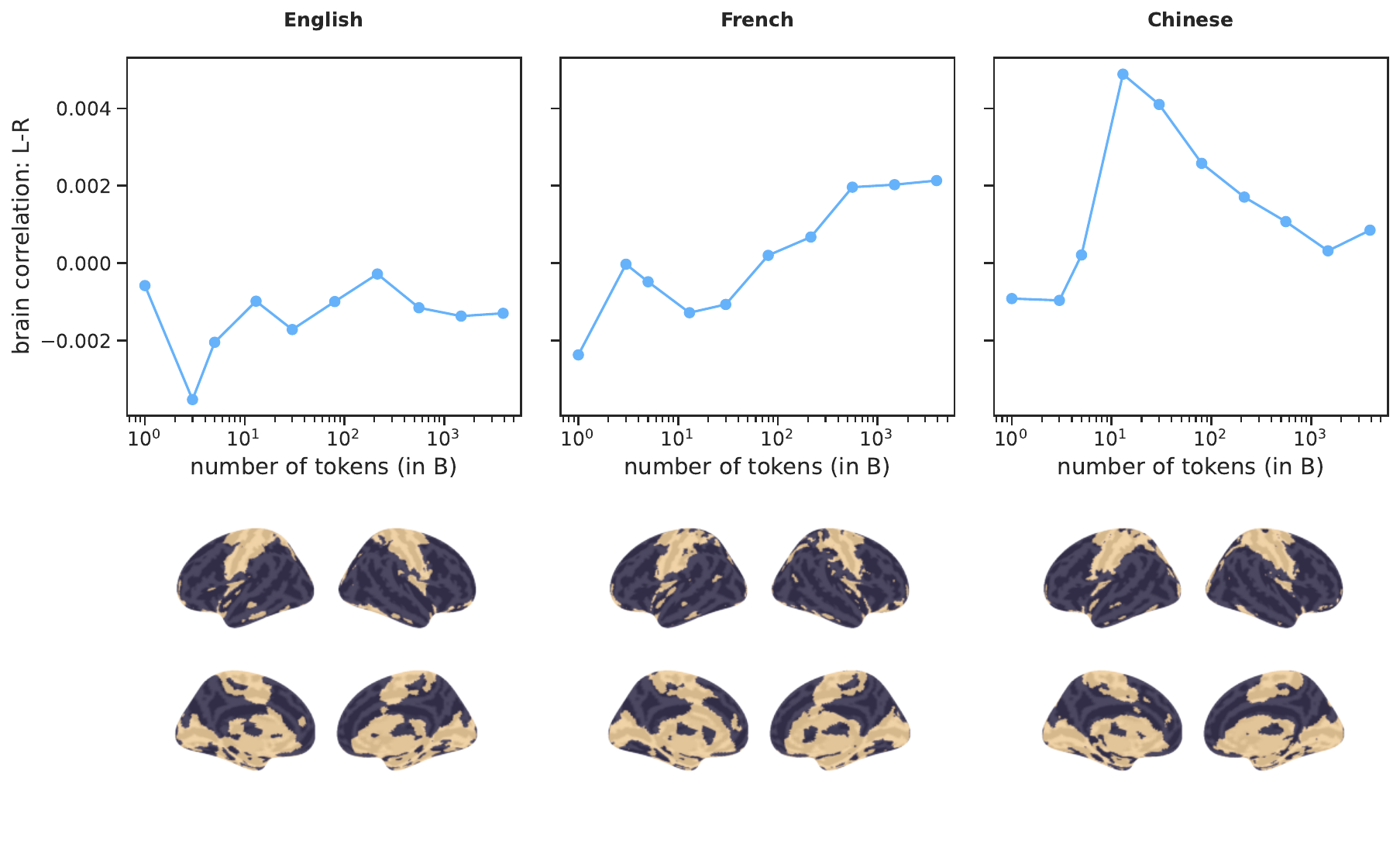}%
\caption{\textbf{Left-right asymmetry computed on the 50\% least reliable voxels.} The voxels are located in regions such as visual or motor cortex. In the case of Chinese, we observe an unexpected spike at the time during training of the discrepancy between whole brain and all the other methods of computing the left-right asymmetry, as seen in Fig.~\ref{fig:olmo-2_l_r_asym_comparison}.
}
\label{fig:olmo-2_l_r_asym_urv}
\end{figure}

\clearpage
\section*{Examples of texts generated by OLMo-2 7B for different training checkpoints}
\label{sec:supp_llm_gen}
This section provides examples of texts generated by the OLMo-2 7B model for the 10 different training checkpoints selected for analysis in the present study, from a given prompt (``Are you'') and a given random seed. Only the first 500 characters of each text are displayed to save space. Each sample starts with a number (1B, 3B, 5B, etc) indicating the number of tokens (in billions) seen by the corresponding checkpoint.\\
All generated texts will be provided in the \texttt{generated\_texts.zip} file available on the GitHub page of the project. Because we noticed some texts contain harmful, toxic, or explicit content, we used a custom Python script running ShieldGemma\footnote{Available at \url{https://huggingface.co/google/shieldgemma-2b}}~\citep{zeng2024shieldgemmagenerativeaicontent} to automatically filter them out.

{\sloppy
\parindent 0pt
\parskip .5pc
\fontsize{9}{9}\selectfont
\ttfamily
\noindent\textbf{---------------------------------------------------------------------------------------- 1B ----------------------------------------------------------------------------------------

Are you is you your. " in what to you, you, or they, in, be. m at, you have out, He was.

", you are, but, and, " In the " per of a of her, you need your these you, in., and at are a to no that your about in the same on it, but. The a, but you need is the the about her it on the the on the The way with the way you a " "The is an " The other with the the the right of the, but that the way of the 1..  This a  This is the the the first of the their. This, and that did the same is no one of some. "a

---------------------------------------------------------------------------------------- 3B ----------------------------------------------------------------------------------------

Are you is you are trying to the people to believe he has to believe they cannot be able to believe their behavior. "What are you here?"" cannot you give you to a young day." He can be a part of a world, like a boy, who is in the same room at a time to avoid getting your eyes in the face to take care of themselves that is not so important. In fact, we must know that if your son will make your own thoughts, you should try and see if you want to make that decision? It is not enough to take a walk 

---------------------------------------------------------------------------------------- 5B ----------------------------------------------------------------------------------------

Are you would go back for the first day to read, you probably don't know what you don't want at your own pace.

And it was really interesting, you could be telling, 'Don't give them some big thumbs,'" she said.

"He's a big stick maker, just like 'Don't know where he's going.' And the things on it, I don't remember where you got off so long. And what it really helped me on was to have the opportunity to read everything we have, and to have an idea that's what you want to be doing, it takes the t

---------------------------------------------------------------------------------------- 13B ----------------------------------------------------------------------------------------

Are you using the new method of the day to help your clients, or are you using the old way?

A few years ago when I was doing some consulting, I was telling people I had worked in the past, and I was telling them I'd been in the industry 12 years, and at that point I'd just kind of left the company. But, I don't remember any of the stories or the details, but it seems like I did get to have the opportunity to go into a variety of things at one point. I got to travel for a long period of time, I 

---------------------------------------------------------------------------------------- 30B ----------------------------------------------------------------------------------------

Are you using the new Google Mail? This is a great way to keep the email conversation in the same place with just a few clicks.

For the uninitiated, Google is a search engine, email service, and a lot more. And this new method is a perfect way to make Google Mail better and less intrusive for our everyday life.

Here we’ll take you through the steps.

  1. Sign up

For our email account, the most efficient way to open Google Mail is through a Google account (see here). Once signed up you will h

---------------------------------------------------------------------------------------- 80B ----------------------------------------------------------------------------------------

Are you so sure your child won't be happy here, you've got to send them somewhere else. Or at least you have to make it so miserable that they'll leave."

She smiled and put her hands on her hips. "Is that what you think your job is here in the school?"

He scrunched up his face. "No, actually it's about getting that kid to behave. He is completely out of control. He's disruptive on the school bus. He throws things at teachers and students."

She smirked at that one. It was one of the funnier pa

---------------------------------------------------------------------------------------- 214B ----------------------------------------------------------------------------------------

Are you using the new Google Play Services SDK to develop your apps, or are you using the older GCM API?
GCM: We use the older GCM.

Do you have a best practice for apps who will use the Google Play Services SDK?
Don’t use it. We had some performance issues and we’re working with Google to make the SDK better. We won’t use it.

What’s the performance issue?
For some apps on Android devices, they run out of memory and have problems. It’s a Google issue, though, not an Ionic issue.

Android is a v

---------------------------------------------------------------------------------------- 558B ----------------------------------------------------------------------------------------

Are you using the new Google Analytics 4 to track your clients, or are you using the normal Google Analytics? How does it compare?

For the complete overview, you are going to want to download the free report which contains all this information. But you need a link to Google in order to do this. This is where our free 3-in-1 Google tool could come in.

  1. In order to download the Google Analytics 4 free report, follow the instructions below:
  2. Click on your profile picture in the top right 

---------------------------------------------------------------------------------------- 1477B ----------------------------------------------------------------------------------------

Are you using the new Google Play Music service to play music you've purchased with the Google Music store? Or are you using the legacy, store? It may be time to reconsider.
Google Music

If you are using the new Google Play Music, you need to upgrade your Google Music. Google is now blocking accounts with older versions of the Google Music app on Google Play from using that version of the app. Why are they doing this? It's pretty simple: to force you to pay for Google Music.

Previously, Google

---------------------------------------------------------------------------------------- 3896B ----------------------------------------------------------------------------------------

Are you using the new tools we made available to you, you've still got the old ones too. Use the new ones. They make it easier.

"Your customers are telling you, 'Give us the good stuff. I don't want this anymore.' "

That means you want people talking about what they're watching. They're talking about the things they're watching because it's good and because you've given them the tools.

It seems like the most efficient way to engage with people is through a combination of things.  You can have

}}

\end{document}